\documentclass[twocolumn,pre,floatfix,superscriptaddress,10pt]{revtex4-2}
\pdfoutput=1

\usepackage[T1]{fontenc}
\usepackage[latin9]{inputenc}
\usepackage{mathrsfs}
\usepackage{amsmath}
\usepackage{amssymb}
\usepackage{graphicx}
\usepackage{graphics}
\usepackage{booktabs}
\usepackage{braket}
\usepackage{color}
\usepackage[svgnames]{xcolor}
\usepackage{comment}
\usepackage{natbib}
\usepackage{dsfont}
\usepackage{bm}
\usepackage{enumitem}
\usepackage{nicefrac}
\usepackage{pgf}
\usepackage{lmodern}

\usepackage{subcaption}
\DeclareMathOperator*{\argmax}{argmax}

\usepackage{mathtools}
\DeclarePairedDelimiter\ceil{\lceil}{\rceil}

\makeatletter

\newcommand{\tX}{\bm{X}}
\newcommand{\tXp}{\bm{X_{\scriptscriptstyle{P}}}}
\newcommand{\tXn}{\bm{X_{\scriptscriptstyle{N}}}}
\newcommand{\ttX}{\bm{\tilde{X}}}

\newcommand{\tetap}{\bm{\eta_{\scriptscriptstyle{P}}}}
\newcommand{\tetan}{\bm{\eta_{\scriptscriptstyle{N}}}}

\newcommand{\tb}{\bm{b}}
\newcommand{\ttb}{\bm{\tilde{b}}}
\newcommand{\np}{n_{\scriptscriptstyle{P}}}
\newcommand{\nn}{n_{\scriptscriptstyle{N}}}
\newcommand{\twp}{w_{\scriptscriptstyle{P}}}
\newcommand{\twn}{w_{\scriptscriptstyle{N}}}
\newcommand{\codevar}[1]{\texttt{#1}}
\newcommand{\msim}{${\sim}$} 
\newcommand{\diag}{\mathrm{diag}}

\usepackage{listings}
\usepackage{hyperref}
\usepackage[capitalise]{cleveref} 

\hypersetup{
  colorlinks=true,
  linkcolor=blue,
  citecolor=ForestGreen,
  urlcolor=DarkOrchid
}

\newlist{questions}{enumerate}{2}
\setlist[questions,1]{label=\textbf{RQ\arabic*.},ref=RQ\arabic*}
\setlist[questions,2]{label=(\alph*),ref=\thequestionsi(\alph*)}

\crefname{listing}{Algorithm}{Algorithms}
\Crefname{listing}{Algorithm}{Algorithms}

\newtheorem{definition}{Definition}

\AtBeginDocument{
  
}

\makeatother

\usepackage{babel}

\begin{document}

\title{Explainable AI using expressive Boolean formulas}

\author{Gili~Rosenberg}
\affiliation{Amazon Quantum Solutions Lab, Seattle, WA 98170, USA}
\thanks{Corresponding author: gilir@amazon.com\\ \\ Fidelity Public Information}

\author{J.~Kyle~Brubaker}
\affiliation{Amazon Quantum Solutions Lab, Seattle, WA 98170, USA}

\author{Martin~J.~A.~Schuetz}
\affiliation{Amazon Quantum Solutions Lab, Seattle, WA 98170, USA}
\affiliation{AWS Center for Quantum Computing, Pasadena, CA 91125, USA}

\author{Grant~Salton}
\affiliation{Amazon Quantum Solutions Lab, Seattle, WA 98170, USA}
\affiliation{AWS Center for Quantum Computing, Pasadena, CA 91125, USA}
\affiliation{California Institute of Technology, Pasadena, CA 91125, USA}

\author{Zhihuai~Zhu}
\affiliation{Amazon Quantum Solutions Lab, Seattle, WA 98170, USA}

\author{Elton~Yechao~Zhu}
\affiliation{Fidelity Center for Applied Technology,
FMR LLC, Boston, MA 02210, USA}

\author{Serdar~Kad{\i}o\u{g}lu}
\affiliation{AI Center of Excellence, FMR LLC, Boston, MA 02210, USA}

\author{Sima~E.~Borujeni}
\affiliation{Fidelity Center for Applied Technology,
FMR LLC, Boston, MA 02210, USA}

\author{Helmut~G.~Katzgraber}
\affiliation{Amazon Quantum Solutions Lab, Seattle, WA 98170, USA}

\date{\today}

\begin{abstract}
We propose and implement an interpretable machine learning classification model for Explainable AI (XAI) based on expressive Boolean formulas. Potential applications include credit scoring and diagnosis of medical conditions. The Boolean formula defines a rule with tunable complexity (or interpretability), according to which input data are classified. Such a formula can include any operator that can be applied to one or more Boolean variables, thus providing higher expressivity compared to more rigid rule-based and tree-based approaches. 
The classifier is trained using native local optimization techniques, efficiently searching the space of feasible formulas. Shallow rules can be determined by fast Integer Linear Programming (ILP) or Quadratic Unconstrained Binary Optimization (QUBO) solvers, potentially powered by special purpose hardware or quantum devices. We combine the expressivity and efficiency of the native local optimizer with the fast operation of these devices by executing non-local moves that optimize over subtrees of the full Boolean formula. 
We provide extensive numerical benchmarking results featuring several baselines on well-known public datasets. Based on the results, we find that the native local rule classifier is generally competitive with the other classifiers. The addition of non-local moves achieves similar results with fewer iterations, and therefore using specialized or quantum hardware could lead to a speedup by fast proposal of non-local moves. 
\end{abstract}

\date{\today}

\maketitle

\section{Introduction}
\label{sec:introduction}

Most of today's machine learning (ML) models are complex, and their inner workings (sometimes with billions of parameters) are difficult to understand and interpret. Yet, in many applications, explainability is desirable or even mandatory due to industry regulations, especially in high-stakes situations (for example, in finance or health care). In situations like these, explainable models can help through increased transparency, with the following additional benefits:\\ 1) explainable models may expose biases, important in the context of ethical and responsible usage of ML, and 2) may be easier to maintain and improve. 

While there exist techniques that attempt to explain black-box ML model decisions \cite{burkart2021survey}, they can be problematic due to their ambiguity, imperfect fidelity, lack of robustness to adversarial attacks, and likely drift from the ground truth \cite{slack2020fooling, lakkaraju2020robust}. Other approaches to explainability focus on constructing interpretable ML models, at times at the price of lower performance \cite{letham2015interpretable, wang2015falling, lakkaraju2016interpretable, ustun2016supersparse, angelino2017learning}. 

In this work we propose an interpretable ML classification model based on \emph{expressive Boolean formulas}. The Boolean formula defines a rule (with tunable complexity and interpretability) according to which input data are classified. Such a formula can include any operator that can be applied to one or more Boolean variables, such as \texttt{And} and \texttt{AtLeast}. This flexibility provides higher expressivity compared to rigid rule-based approaches, potentially resulting in improved performance and interpretability. 

Quantum computers might offer speedups for solving hard optimization problems in the fullness of time \cite{zahedinejad2017combinatorial, sanders2020compilation}. In the near term, specialized, classical hardware has been developed for speeding up ML \cite{reuther2019survey, bavikadi2022survey} and optimization workloads \cite{aramon2019physics, mohseni2022ising}. Such devices, classical or quantum, tend to have a limited scope of applicability (i.e., areas of potential advantage). Moreover, native optimization---i.e., the solving of optimization problems in their natural representation, promises to be more efficient \cite{valiante2021computational}, but often requires a custom optimizer that cannot easily take advantage of specialized hardware. For the explainability problem, we develop a native optimization algorithm that utilizes specialized hardware to efficiently solve subproblems, thereby combining the advantages of both techniques---specialized hardware and native optimization. 

The main contributions of this paper are:
\begin{itemize}
    \item Improved and expanded Integer Linear Programming (ILP) formulations and respective Quadratic Unconstrained Binary Optimization (QUBO) formulations for finding depth-one rules.
    \item A native local solver for determining expressive Boolean formulas.
    \item The addition of non-local moves, powered by the above ILP/QUBO formulations (or potentially other formulations).
\end{itemize}

The main findings are: 
\begin{itemize}
    \item Expressive Boolean formulas provide a more compact representation than decision trees and conjunctive normal form (CNF) rules for various examples. 
    \item Parameterized operators such as \texttt{AtLeast} are more expressive than non-parametrized operators such as \texttt{Or}. 
    \item The native local rule classifier is competitive with well-known alternatives considered in this work. 
    \item The addition of non-local moves achieves similar results with
fewer iterations, and therefore using specialized or quantum hardware could lead to a speedup by fast proposal of non-local moves.
\end{itemize}

This paper is structured as follows. In \cref{sec:related_work}, we review related work, and in \cref{sec:preliminaries} we provide the problem definition, introduce expressive Boolean formulas, and motivate their usage for explainable artificial intelligence (XAI). In \cref{sec:native_optimization}, we lay out the training of the classifier using a native local optimizer, including the idea and implementation of non-local moves. In \cref{sec:ilp_qubo_formulations} we formulate the problem of finding optimal depth-one rules as ILP and QUBO problems. In \cref{sec:results}, we present and discuss our results, and in \cref{sec:conclusions} we present our conclusions and discuss future directions.

\section{Related work} 
\label{sec:related_work}

Broadly speaking, there are two prevalent approaches to XAI which we briefly (briefly reviewed below):

\textbf{Post-hoc explanation of black-box models (Explainable ML)} --
Many state of the art ML models, particularly in Deep Learning (DL), are huge---consisting of a large number of weights and biases, recently surpassing a trillion parameters \cite{fedus2022switch}. These DL models are, by nature, difficult to decipher. The most common XAI approaches for these models provide post-hoc explanation of black-box model decisions. These approaches are typically model agnostic and can be applied to arbitrarily complex models, such as the ones commonly used in DL. 

\textit{Local explanation methods} apply local approximations to groups of instances, such as LIME \cite{ribeiro2016should} and SHAP \cite{lundberg2017unified}, or to different feature sets, such as MUSE \cite{lakkaraju2019faithful}. Such methods often suffer from a lack of robustness and can typically be fooled easily by adversarial attacks
\cite{slack2020fooling, lakkaraju2020robust}.	

\textit{Global explanation methods} aim to mimic black-box models using interpretable models \cite{craven1995extracting, bastani2017interpreting}. These methods benefit from the easy availability of additional data, by querying the black-box model. However, there is an issue of ambiguity---different interpretable models might yield the same high fidelity to the black-box model. Furthermore, since it is generally impossible to achieve perfect fidelity, the resulting model can drift even farther from the ground truth than the original black-box model. 

\textbf{Training interpretable models (Interpretable ML)} --
The definition of what constitutes an interpretable model is domain-specific and potentially user specific. Nevertheless, many interpretable models have been studied. As a few selected examples, see rule lists \cite{letham2015interpretable}, falling-rule lists \cite{wang2015falling}, decision sets \cite{lakkaraju2016interpretable}, scoring systems \cite{ustun2016supersparse}, and the more well-known decision trees and linear regression. There is an existing body of work on using specific Boolean formulas as ML models. For example, learning conjunctive/disjunctive normal form (CNF/DNF) rules using a MaxSAT (Maximum Satisfiability) solver \cite{malioutov2018mlic, ghosh2019imli}, an ILP solver \cite{su2015interpretable, wang2015learning, lawless2021interpretable}, or via LP-relaxation \cite{malioutov2017learning}. 

There is a common belief that interpretable models yield less accurate results than more complex black-box models. However, in numerous cases, interpretable models have been shown to yield comparable results to black-box models \cite{rudin2019stop}, which is often the case for structured data with naturally meaningful features. Nonetheless, there may be cases in which interpretable models in fact do yield worse results. In those cases, a loss of accuracy might be a worthwhile concession in exchange for the additional trust that comes with an interpretable model. 

Interpretable classifiers, including the classifiers described in this work can be used as stand-alone interpretable models, or they can be used to explain black-box models. The main difference is in the origin of the labels---in the former they would be the ground truth, whereas in the latter they would be the output of the black-box model.

\section{Preliminaries}
\label{sec:preliminaries}

In this section we introduce our problem definition, the objective functions, expressive Boolean formulas and motivate their usage for XAI. 

\subsection{Problem definition}

We start by stating our problem definition in words:

\begin{definition}[\textbf{Rule Optimization Problem (ROP)}]
Given a binary feature matrix $\bm{X}$, and a binary label vector $\bm{y}$, the goal of the Rule Optimization Problem (ROP) is to find the optimum rule $R^{*}$, that balances the score $S$ of the rule $R$ on classifying the data, and the complexity of the rule $C$, which is given by the total number of features and operators in $R$. The complexity might in addition be bounded by a parameter $C'$. 
\label{def:problem}
\end{definition}

Mathematically, our optimization problem can be stated at a high-level as:
\begin{alignat}{2}
\label{eq:rule_objective_function}
&R^{*} = \argmax_R[S(R(\bm{X}), \bm{y}) - \lambda \, C(R)] \\
&\text{s.t.} \quad C(R) \leq C', \notag
\end{alignat}
where $R$ is any valid rule, $R^{*}$ is the optimized rule, $S$ is the score, $C$ is the complexity, $\bm{X}$ is a binary matrix containing the input data, $\bm{y}$ is a binary vector containing the corresponding labels, and $\lambda$ is a real number that controls the tradeoff between the score and the complexity. The solution of this problem is a single rule. Note that our problem definition is flexible to accommodate different design decisions which are described in more detail below.

\subsection{Objectives}

In this problem we have two competing objectives: maximizing the performance, and minimizing the complexity. The performance is measured by a given metric that yields a score $S$. 

This problem could be solved by a multi-objective optimization solver, but we leave that for future work. Instead, we adopt two common practices in optimization: 
\begin{itemize}
\item \textit{Combining multiple objectives into one} -- introducing a new parameter $\lambda \geq 0$ that controls the relative importance of the complexity (and, therefore, interpretability). The parameter $\lambda$ quantifies the drop in the score we are willing to accept, in order to decrease the complexity by one. Higher values of $\lambda$ generally result in less complex models. We then solve a single-objective optimization problem with the objective function  $S - \lambda C$, which combines both objectives into one hybrid objective controlled by the (use-specific) parameter $\lambda$. 
\item \textit{Constraining one of the objectives} -- introducing the maximum allowed complexity, $C'$ (also referred to as \codevar{max\_complexity}), and then varying $C'$ to achieve the desired result.
\end{itemize} 

Normally, formulations include only one of these methods. However, we choose to include both in our formulation, for the following reasoning. The former method does not provide guidance on how to set $\lambda$, while the latter method provides tight control over the complexity of the rule, at the price of including an additional constraint. In principle, we prefer the tight control provided by setting $C'$, since adding just one constraint is not a prohibitive price to pay. However, note that if $\lambda=0$, solutions that have an equal score but different complexity have the same objective function value. In reality, in most use cases we expect that the lower complexity solution would be preferred. In order to indicate this preference to the solver, we recommend setting $\lambda$ to a small, nonzero value. Regardless, in our implementation users can set $C'$, $\lambda$, both, or neither of them. 

Without loss of generality, in this work we mainly use \emph{balanced accuracy} as the performance metric. Here, $S$ is equal to the mean of the accuracy of predicting each of the two classes:
\begin{equation}
S = \frac12 \left( \frac{TP}{TP+FN} + \frac{TN}{TN+FP} \right),
\end{equation}
where $TP$ is the number of true positives, $FN$ is the number of false negatives, and similarly for $FP$ and $TN$. Generalizations to alternative metrics are straightforward. For balanced datasets, balanced accuracy reduces to regular accuracy. The motivation for using this metric is that many datasets of interest are not well balanced. 

A common use case is to explore the score vs. complexity tradeoff by varying $C'$ or $\lambda$ over a range of values, producing a series of rules in the score-complexity space, as close as possible to the Pareto frontier. 
\begin{figure}[htb]
\includegraphics[width=0.4 \textwidth]{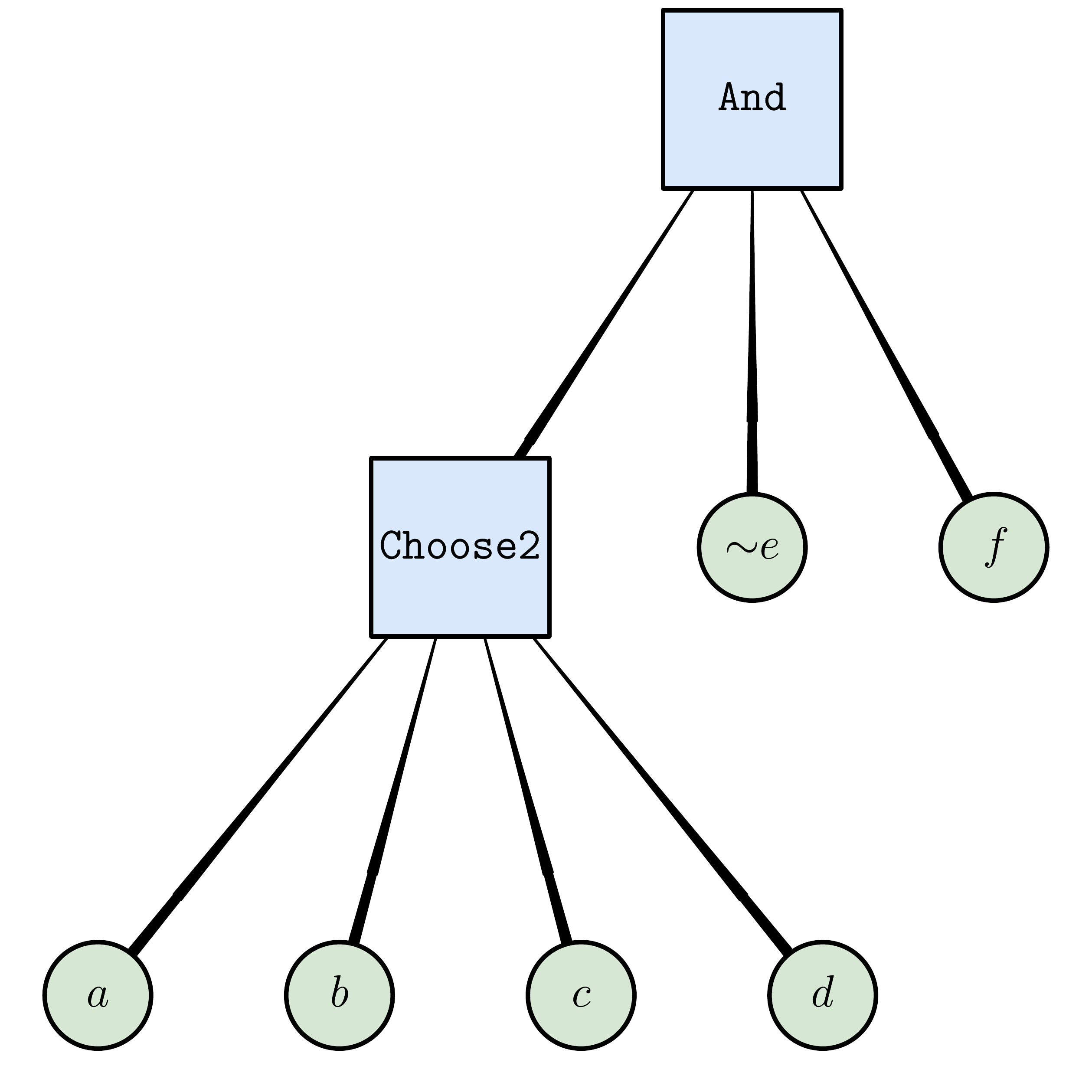}
\caption{A simple expressive Boolean formula. This formula contains six literals and two operators, has a depth of two, and a complexity of eight. It can be stated also as \texttt{And(Choose2($a, b, c, d$)$,{\sim}e, f$)}.}
\label{fig:boolean_formula}
\end{figure}

\subsection{Rules as expressive Boolean formulas}

\begin{figure*}[htb]
\includegraphics[width=0.9 \linewidth]{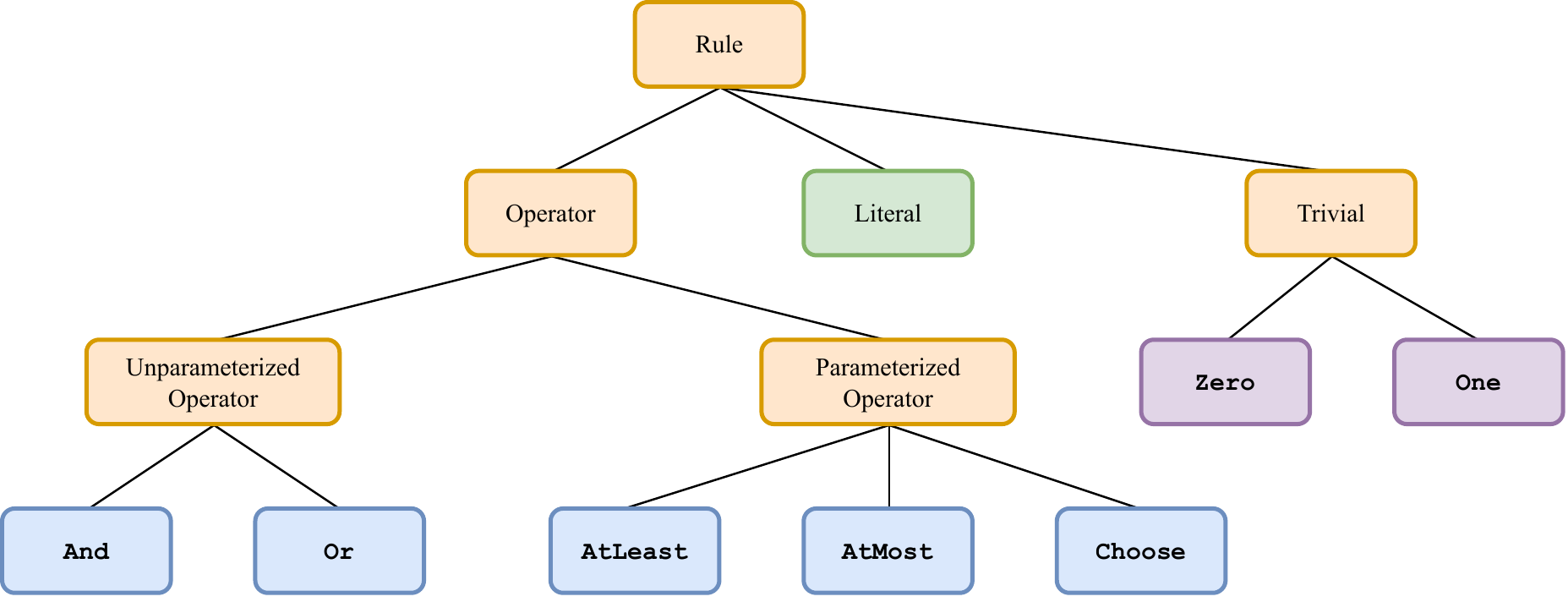}
\caption{A class diagram showing the hierarchy of classes that we use to define expressive Boolean formulas in this work. The leaves in this figure are concrete classes, the rest are abstract classes---not meant to be instantiated, but rather subclassed. The operator classes we have included in this work are divided into two groups: unparametrized operators and parameterized operators. The trivial classes are trivial rules that return zero always (\texttt{Zero}) or one always (\texttt{One}). Literals and operators can optionally be negated. }
\label{fig:rule_class_diagram}
\end{figure*}

The ROP (see \cref{def:problem}) can be solved over different rule definitions. In this work, we define the the rules $R$ to be Boolean formulas, and in particular expressive Boolean formulas. An \emph{expressive Boolean formula} (henceforth, a ``formula''), as we define it here, consists of literals and operators. Literals are variables $f_i$ or negated variables ${\sim}f_i$. Operators are operations that are performed on two or more literals, such as \texttt{And($f_0, f_1, \mathord\sim f_2$)} and \texttt{Or($\mathord\sim f_0, f_1$)}. Some operators are parameterized, for example \texttt{AtLeast2($f_0, f_1, f_2$)}, which would return true only if at least two of the literals are true. Operators can optionally be negated as well. For simplicity, we consider negation to be a property of the respective literal or operator rather than being represented by a \texttt{Not} operator. See \cref{fig:rule_class_diagram} for the operators we have included in this work. The definitions and implementation we have used are flexible and modular---additional operators could be added (such as \texttt{AllEqual} or \texttt{Xor}), or some could be removed. 

The inclusion of operators like \texttt{AtLeast} is motivated by the idea of (highly interpretable) checklists, such as a list of medical symptoms that signify a particular condition. It is conceivable that a decision would be made using a checklist of symptoms, of which a minimum number would have to be present for a positive diagnosis. Another example is a bank trying to decide whether or not to provide credit to a customer. 

It is convenient to visualize formulas as directed graphs (see \cref{fig:boolean_formula}). The leaves in the graph are the literals which are connected with directed edges to the operator operating on them. To improve readability, we avoid crossovers by including a separate node for each literal, even if that literal appears in multiple places in the formula. Formally, this graph is a directed rooted tree. Evaluating a formula on given values of the variables can be done by starting at the leaves, substituting the variable values, and then applying the operators until one reaches the top of the tree, referred to as the \textit{root}. 

We define the \textit{depth} of a formula as the longest path from the root to any leaf (literal). For example, the formula in \cref{fig:boolean_formula} has a depth of two. We define the \textit{complexity} as the total number of nodes in the tree, i.e., the total number of literals and operators in the formula. That same formula has a complexity of eight. This definition is motivated by the intuitive idea that adding literals or operators generally makes a given formula less interpretable. In this study, we are concerned with maximizing interpretability, which we shall do by minimizing complexity.

\subsection{Motivation}
\label{sec:motivation}

In this section we motivate our work by illustrating with a few simple examples how rigid-rule based classifiers and decision trees can require unreasonably complex models for simple rules. 

\begin{figure*}[htb]
\includegraphics[width=1.0 \linewidth]{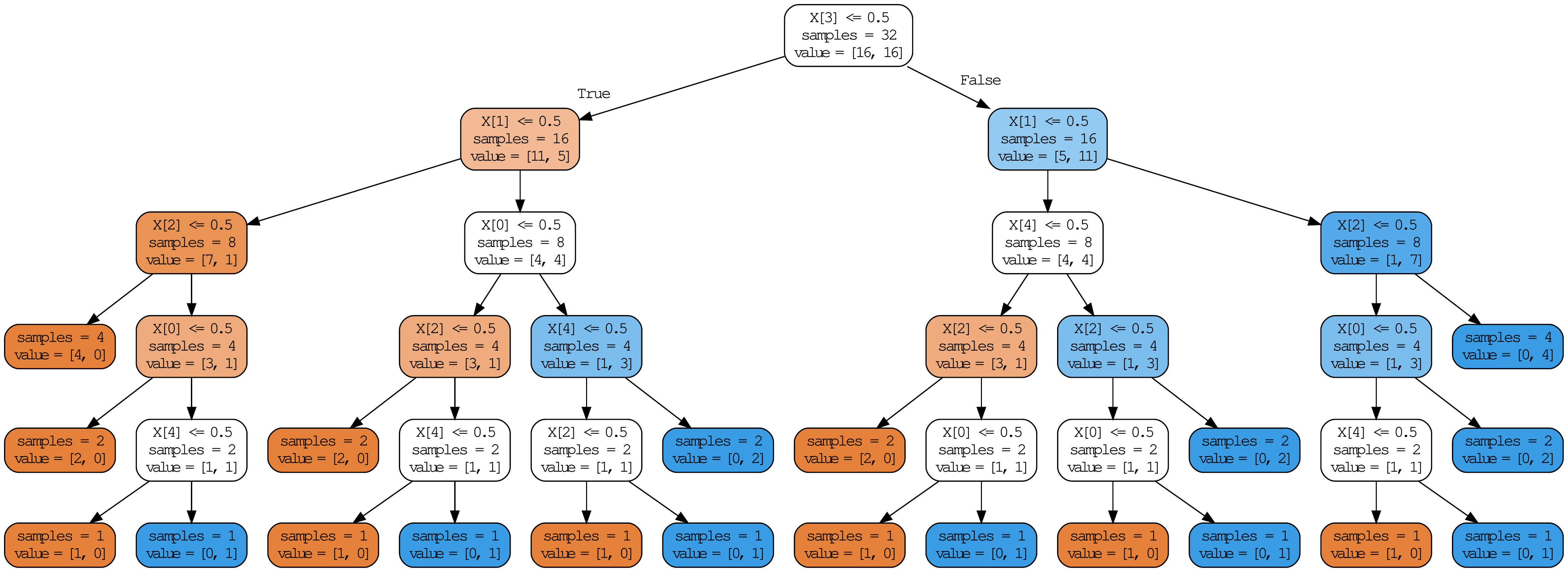}
\caption{Simple example rule that yields a complex decision tree. The optimal rule for this dataset can be stated as \texttt{AtLeast3($f_0, \dots, f_4$)}, yet the decision tree trained on this dataset has 19 split nodes and is not easily interpretable.}
\label{fig:decision_tree_atleast_3}
\end{figure*}

Shallow decision trees are generally considered highly interpretable and can be trained fairly efficiently. However, it is easy to construct simple datasets that require very deep decision trees to achieve high accuracy. For example, consider a dataset with five binary features in which data rows are labelled as true only if at least three of the features are true. This is a simple rule that can be stated as {\texttt{AtLeast3($f_0, \dots, f_4$)}}. However, training a decision tree on this dataset results in a large tree with 19 split nodes (see \cref{fig:decision_tree_atleast_3}). Despite encoding a simple rule, this decision tree is deep and difficult to interpret. 

The prevalence of methods of finding optimal CNF (or equivalently, DNF) rules using MaxSAT solvers \cite{malioutov2018mlic, ghosh2019imli} or ILP solvers \cite{su2015interpretable, wang2015learning, lawless2021interpretable} suggests that one might use such a formula as the rule for the classifier. However, in this case as well, it is easy to construct simple datasets that require complicated rules. Consider the example above---the rule {\texttt{AtLeast3($f_0, \dots, f_4$)}} requires a CNF rule with 11 literals, 13 clauses, and a rule length of 29 (number of literals in all clauses). 

\begin{figure}[htb]
\includegraphics[width=1.0 \columnwidth]{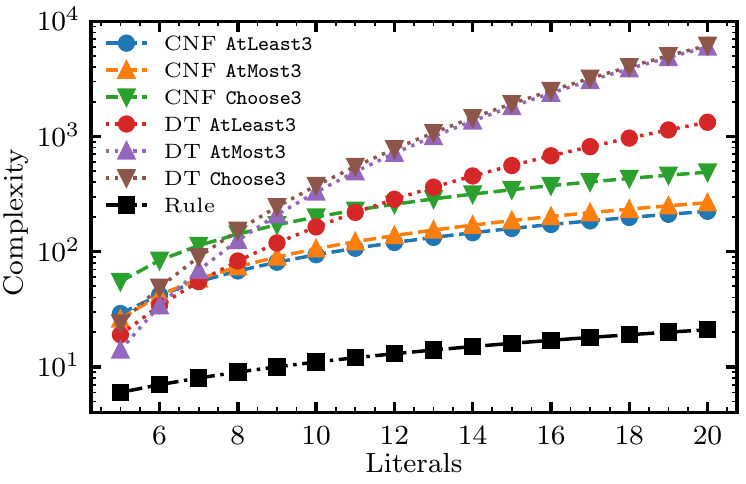}
\caption{A comparison of the complexity required to represent rules of the form \texttt{AtLeast3}, \texttt{AtMost3}, and \texttt{Choose3}, varying the number of literals included under the operator. ``CNF'' is a CNF formula encoded via sequential counters \cite{sinz2005towards}, as implemented in \textsc{pysat} \cite{pysat}. ``DT'' is a decision tree, as implemented in \textsc{scikit-learn} \cite{scikit-learn}. ``Rule'' is an expressive Boolean formula as defined in this paper. The complexity of a decision tree is defined as the number of decision nodes. The complexity of a CNF formula is defined as the total number of literals that appear in the CNF formula (including repetitions). The complexity of expressive Boolean formulas is defined as the total number of operators and literals (including repetitions), so in this case it is equal to the number of literals plus one.}
\label{fig:encoding_complexity}
\end{figure}

\cref{fig:encoding_complexity} shows several examples in which decision trees and CNF rules require a complicated representation for simple rules. The complexity $C$ of a decision tree is defined here as the number of decision nodes. The complexity of a CNF formula is defined (conservatively) as the total number of literals that appear in the CNF formula (including repetitions). For CNF rules, we also tried other encodings besides the one indicated in the table (sorting networks \cite{batcher1968sorting}, cardinality networks \cite{asin2009cardinality}, totalizer \cite{bailleux2003efficient}, modulo totalizer \cite{ogawa2013modulo}, and modulo totalizer for $k$-cardinality \cite{morgado2014mscg}), all of which produced more complex formulas for this data. One can see that the decision tree encodings (``DT'') are the least efficient, followed by the CNF encodings (``CNF''), and finally expressive Boolean formulas (``Rule'') are by far the most efficient at encoding these rules.

\section{The components of the solver}
\label{sec:native_optimization}

In this work we propose to determine optimized expressive Boolean formulas via native local and non-local optimization. Here, ``native'' refers to optimization in the natural search space for the problem. That is in contrast to reformulating the problem in a fixed-format such as MaxSAT, ILP, or QUBO, which would be difficult (if not impossible), and often requires searching a much larger space. A natural search space for this problem is the space of valid expressive Boolean formulas. Next, ``local'' refers to the exploration of the search space via stochastic local search, i.e., by performing a series of moves that make relatively small changes to the current configuration \cite{hoos2004stochastic}. ``Non-local'' optimization refers to exploration of the search space via moves that change a larger part of the solution, a form of large neighborhood search \cite{pisinger2019large}. Below we describe the native local optimizer in detail, and then we explain the idea of non-local moves and our specific implementation thereof.

In order to define a native local solver, we must define several components: the constraints and search space, how to generate initial rules to start the optimization, the allowed local moves, how to evaluate proposed rules, and the way the solver works. Below we provide information on our choice for each of these components. However, other choices are possible for each of the components and might provide better results.

\subsection{Constraints and search space}

This problem can be posed as an unconstrained optimization problem. In order to do so, we must define the search space such that all candidate solutions are feasible. This search space is infinite if $C'$ is not set. However, in this case the solver focuses only on a small fraction of the search space due to the regularizing effect of $\lambda$. To see this, note that $0 \leq S \leq 1$, and $C \geq 1$. Thus, for a given value of $\lambda$, rules with $C > 1 / \lambda$ yield a negative objective function value, which is surely exceeded by at least one rule. Therefore, these rules would be avoided by the solver, or at least de-emphasized. 

In principle, rules could involve no literals (``trivial rules''), a single literal, an operator that operates on two or more literals, and any nested combination of operators and literals. In practice, trivial rules and single-literal rules can be quickly checked exhaustively, and for any reasonably complicated dataset would not provide high enough accuracy to be of interest. Therefore, we simplify our solver by excluding such rules. In order to check our assumptions (and as a useful baseline), our experiments include the results provided by the optimal single literal (feature) rule, as well as the optimal trivial rule (always one or always zero).

For parametrized operators, we constrain the search space such that it only includes sensible choices of the parameters. Namely, for \texttt{AtMost}, \texttt{AtLeast}, and \texttt{Choose}, we require that $k$ is non-negative, and is no larger than the number of literals under the operator. These constraints are fulfilled by construction, by picking initial solutions and proposing moves that take them into account.

\subsection{Generating initial rules}

Initial rules are constructed by choosing between two and $C'-1$ literals randomly (without replacement). Literals are chosen for negation via a coin flip. Once the literals have been chosen, an operator and valid parameter (if the operator chosen is parametrized) are selected randomly. These generated rules are of depth-one---additional depth is explored by subsequent moves.

\subsection{Generating local moves}

Feasible local moves are found by \textit{rejection sampling} as follows.  A node (literal or operator) in the current rule is chosen at random, and a move type is chosen by cycling over the respective move types for the chosen literal/operator. Next, a random move of the chosen type is drawn. If the random move is invalid, the process is restarted, until a valid move is found (see \cref{algo:propose_local_move}). Typically only a few iterations are needed, at most. 

\begin{figure*}
\begin{center}
\noindent\begin{minipage}{.7\linewidth}
\lstset{caption={The function that proposes local moves---\texttt{propose\_local\_move()}. This function selects a node (an operator or a literal) randomly, while cycling through the move types, and then attempts to find a valid local move for that node and move type. The process is re-started if needed until a valid local move is found, a process which is referred to as ``rejection sampling.''}}
\begin{lstlisting}[language=Python, label={algo:propose_local_move}]
literal_move_types = cycle({"remove_literal",
                            "expand_literal_to_operator",
                            "swap_literal"})
operator_move_types = cycle({"remove_operator",
                             "add_literal",
                             "swap_operator"})

def propose_local_move(current_rule):
    all_operators_and_literals = current_rule.flatten()

    proposed_move = None
    while proposed_move is None:
        target = random.choice(all_operators_and_literals)
        
        if isinstance(target, Literal):
            move_type = next(literal_move_types)
        else: 
            move_type = next(operator_move_types)

        proposed_move = get_random_move(move_type, target)

    return proposed_move
\end{lstlisting}
\end{minipage}
\end{center}
\end{figure*}

\begin{figure*}[htb]
  \begin{minipage}{7cm}
	\begin{subfigure}[t]{7cm}
		\centering
		\includegraphics[width=7cm]{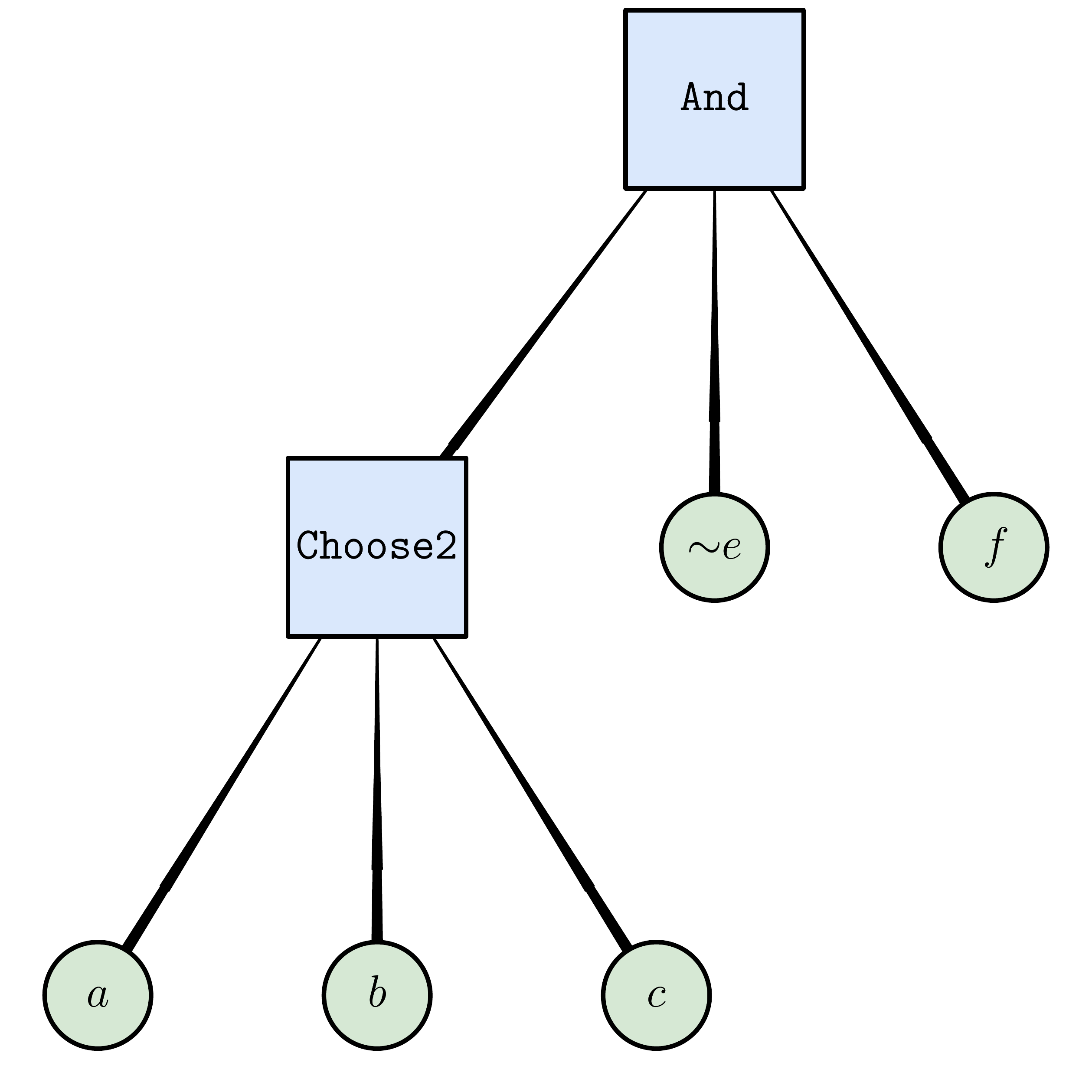}
		\caption{Remove literal $d$.}\label{fig:local_moves_remove_literal}
	\end{subfigure}\par
 	\begin{subfigure}[t]{7cm}
		\centering
		\includegraphics[width=7cm]{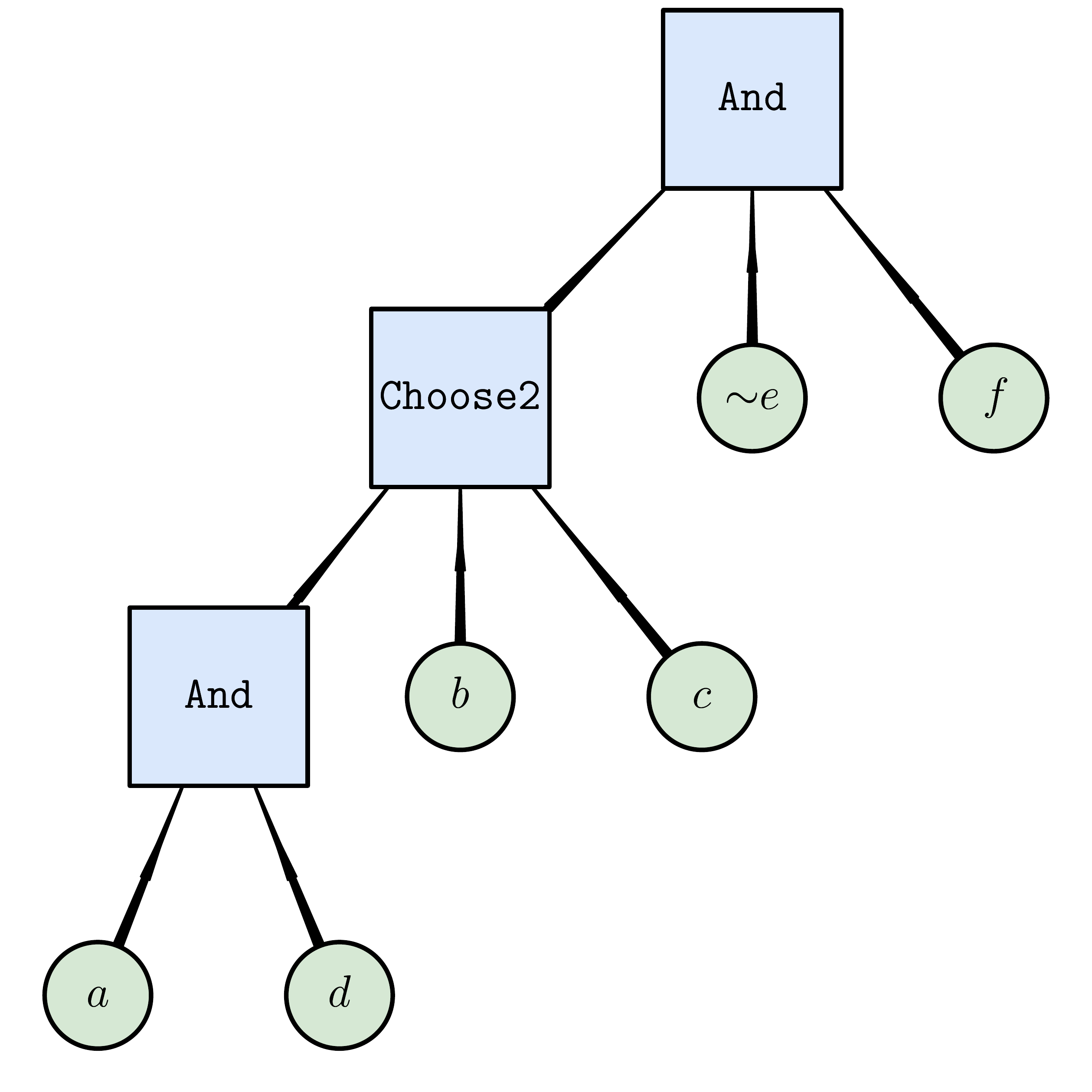}
		\caption{Expand literal $a$ to operator \texttt{And}.}\label{fig:local_moves_expand_literal}
	\end{subfigure}\par
	\begin{subfigure}[t]{7cm}
		\centering
		\includegraphics[width=7cm]{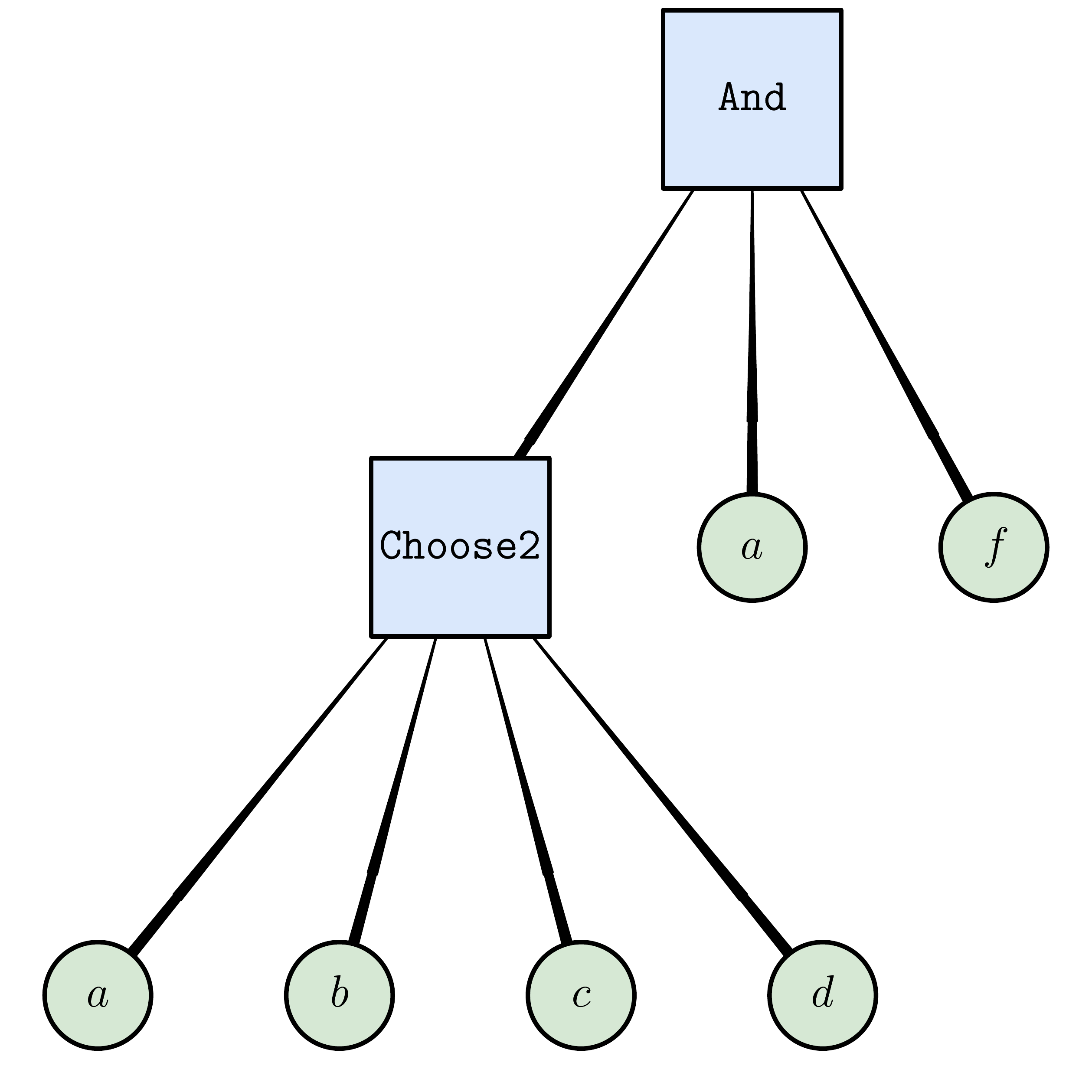}
		\caption{Swap literal $\mathord\sim e$ for $a$.}\label{fig:local_moves_swap_literal}
	\end{subfigure}
  \end{minipage}
  \hspace{2cm}
  \begin{minipage}{7cm}
	\begin{subfigure}[t]{7cm}
		\centering
		\includegraphics[width=7cm]{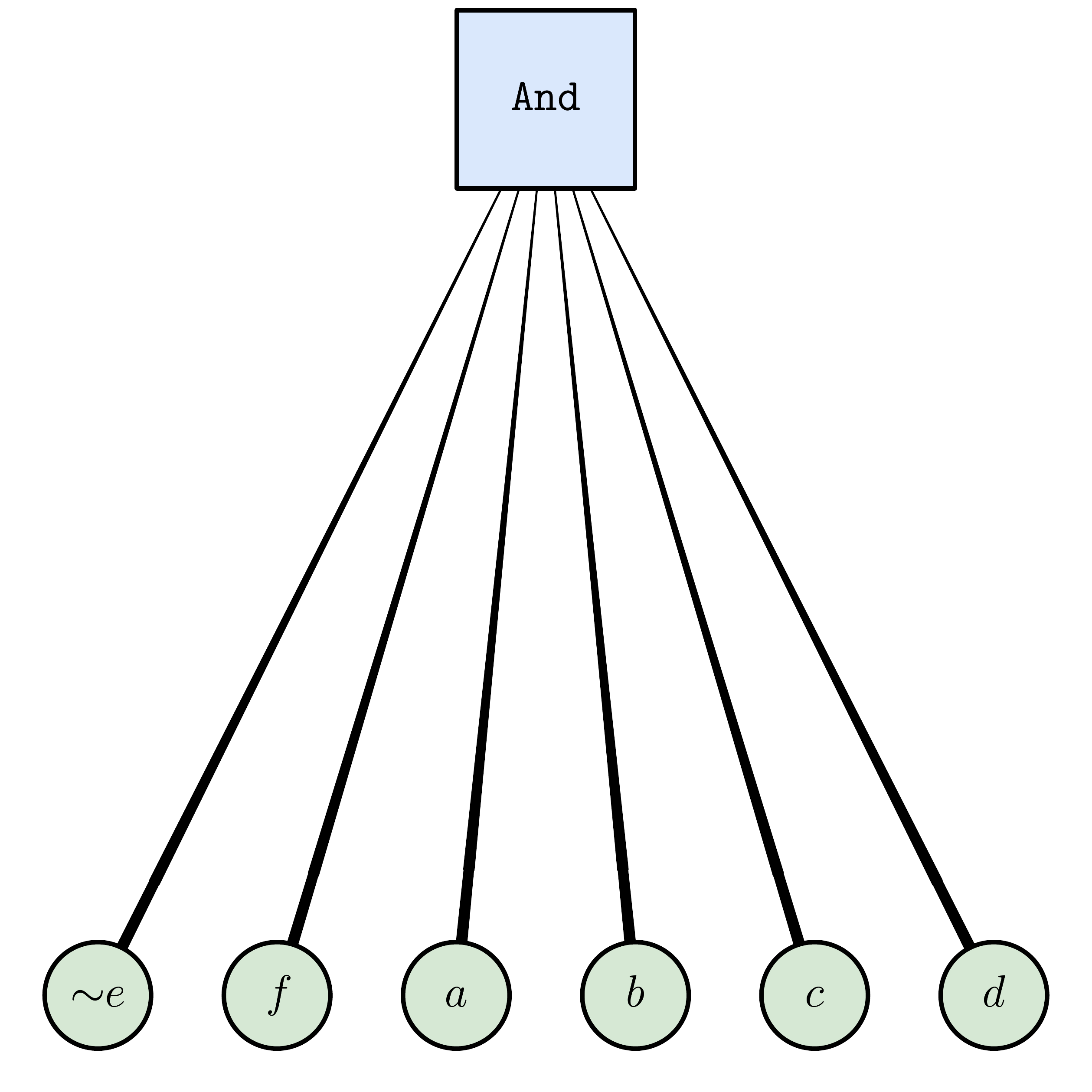}
		\caption{Remove operator \texttt{Choose2}.}\label{fig:local_moves_remove_operator}
	\end{subfigure}\par
	\begin{subfigure}[t]{7cm}
		\centering
		\includegraphics[width=7cm]{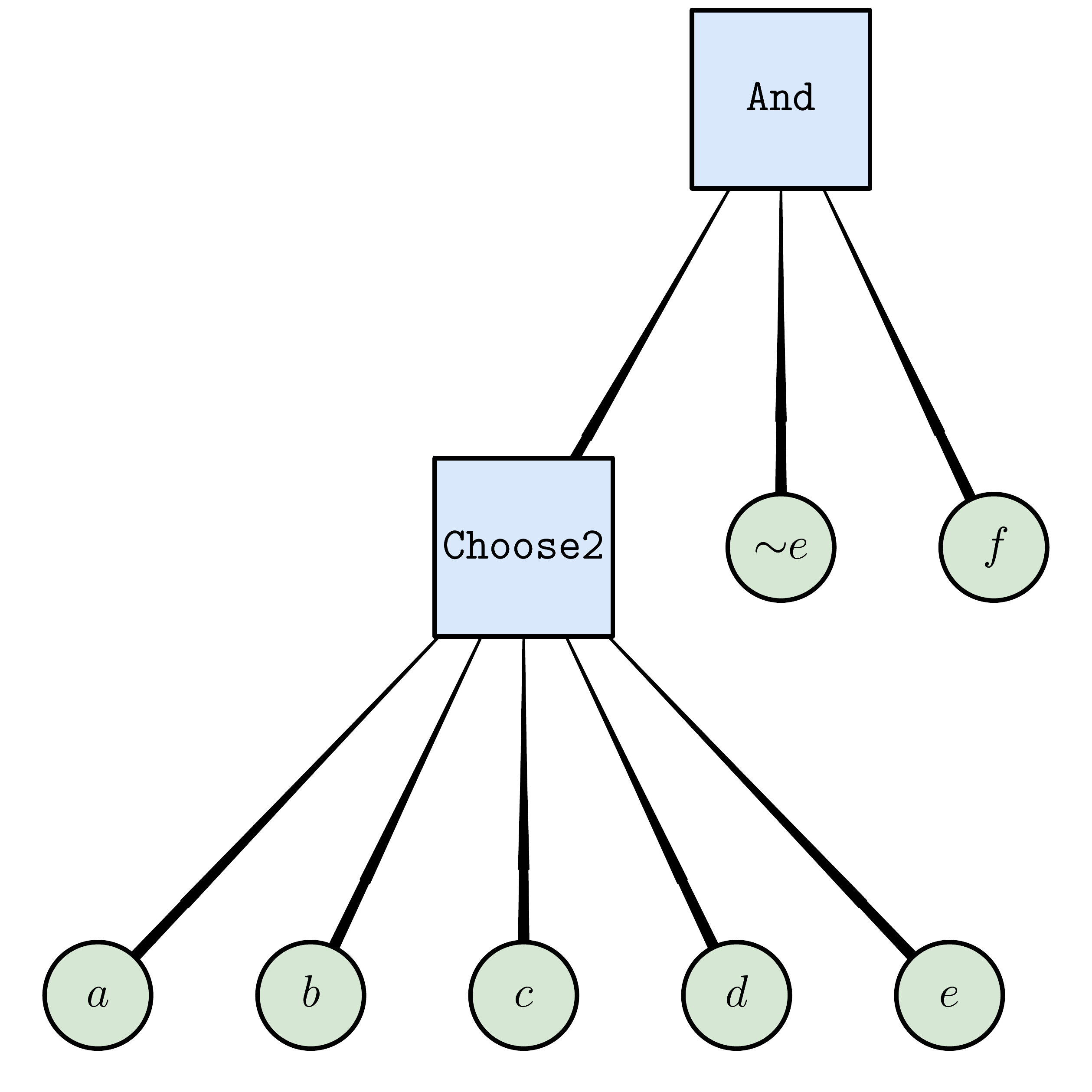}
		\caption{Add literal $e$ to operator \texttt{Choose2}.}\label{fig:local_moves_add_literal}
	\end{subfigure}\par
 	\begin{subfigure}[t]{7cm}
		\centering
		\includegraphics[width=7cm]{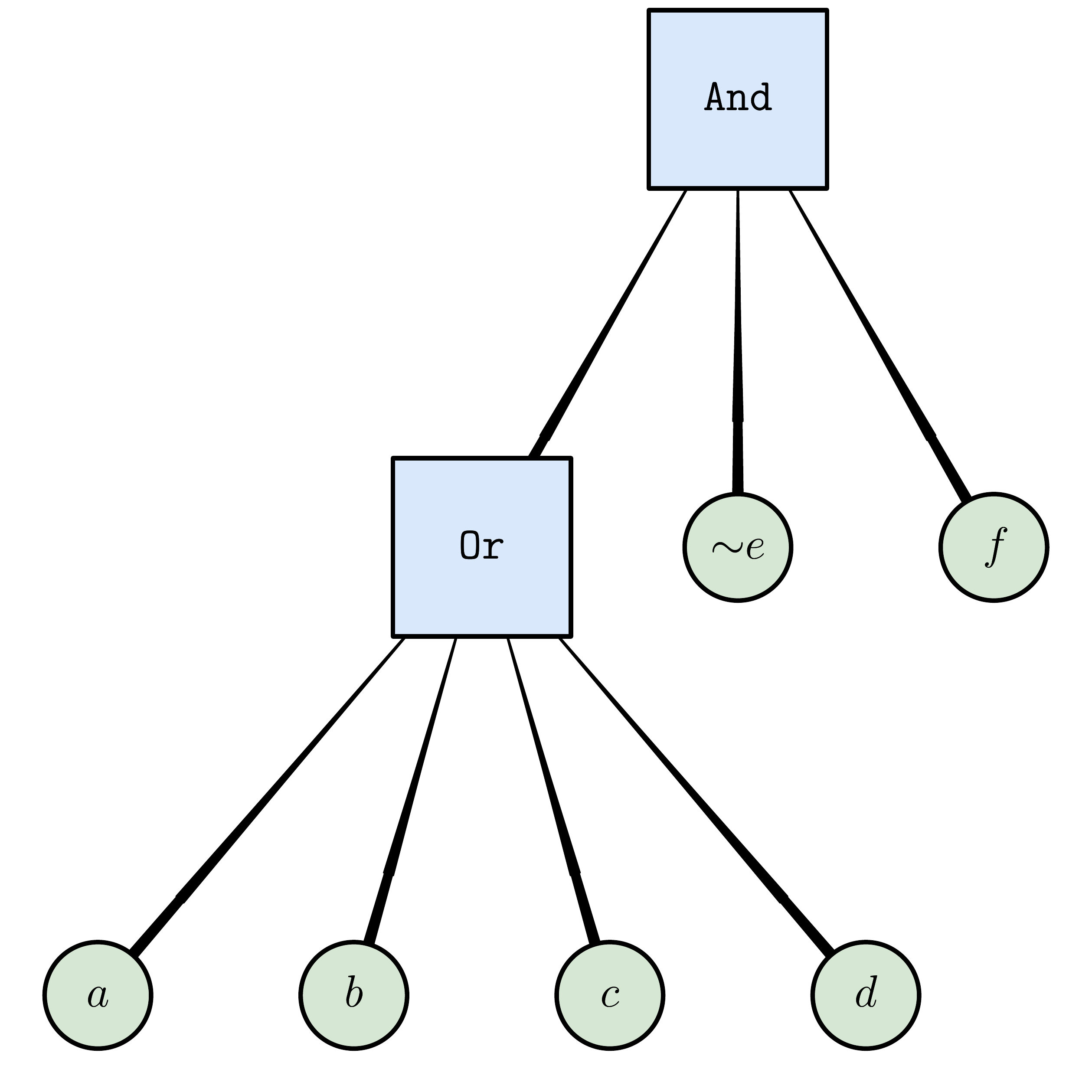}
		\caption{Swap operator \texttt{Choose2} for \texttt{Or}.}\label{fig:local_moves_swap_operator}
	\end{subfigure}
  \end{minipage}
  \caption{Local move types. Moves on literals are shown in (a)--(c) and moves on operators in (d)--(f). All moves are relative to the initial rule shown in \cref{fig:boolean_formula}.}
  \label{fig:local_moves}
\end{figure*}

The literal move types we have implemented are [see \cref{fig:local_moves}(a)--(c)]: 
\begin{itemize}
\item \textit{Remove literal} -- removes the chosen literal but only if the parent operator would not end up with fewer than two subrules. If the parent is a parametrized operator, adjusts the parameter down (if needed) such that it remains valid after the removal of the chosen literal. 
\item \textit{Expand literal to operator} -- expands a chosen literal to an operator, moving a randomly chosen sibling literal to that new operator. Proceeds only if the parent operator includes at least one more literal. If the parent is a parametrized operator, adjusts the parameter down (if needed) such that it remains valid after the removal of the chosen literal and the sibling literal. 
\item \textit{Swap literal} -- replaces the chosen literal with a random literal that is either the negation of the current literal, or is a (possibly negated) literal that is not already included under the parent operator. 
\end{itemize}

The operator move types we have implemented are [see \cref{fig:local_moves}(d)--(f)]:
\begin{itemize}
\item \textit{Remove operator} -- removes an operator and any operators and literals under it. Only proceeds if the operator has a parent (i.e., it is not the root), and if the parent operator has at least three subrules, such that the rule is still valid after the move has been applied. 
\item \textit{Add literal to operator} -- adds a random literal (possibly negated) to a given operator, but only if that variable is not already included in the parent operator. 
\item \textit{Swap operator} -- swaps an operator for a randomly-selected operator and a randomly-selected parameter (if the new operator is parametrized). Proceeds only if the new operator type is different, or if the parameter is different. 
\end{itemize}

\subsection{Evaluation of rules}

The ``goodness'' of a rule is defined by the objective function in \cref{eq:rule_objective_function}, i.e., as $S-\lambda C$. In order to calculate the objective function, we must evaluate the rule with respect to the given data rows. The evaluation yields a Boolean prediction for each data row. Evaluating the metric on the predictions and respective labels results in a score $S$. The complexity of a rule $C$ is a function only of the rule's structure, and does not depend on the inputs. Therefore, the computational complexity of calculating the objective function is dominated by the rule evaluation, which is linear in both the number of samples and the rule complexity.

The evaluation of literals is immediate, because the result is simply equal to the respective feature (data column), or the complement of that feature. The evaluation of operators is done by evaluating all the subrules, and then applying the operator to the result. Evaluation is usually done on multiple rows at once. For this reason, it is far more efficient to implement evaluation using vectorization, as we have done. In fact, for large datasets, it might be beneficial to parallelize this calculation, because it is trivially parallelizable (over the data rows). In addition, evaluation in the context of a local solver could likely be made far more efficient by memoization---i.e., storing the already evaluated subrules so that they can be looked-up rather than re-evaluated. 

\subsection{The native local solver}
\label{sec:local_solver}

\begin{figure*}[htb]
  \begin{minipage}{\textwidth}
	\begin{subfigure}[t]{0.5\textwidth}
	    \centering
	    \includegraphics{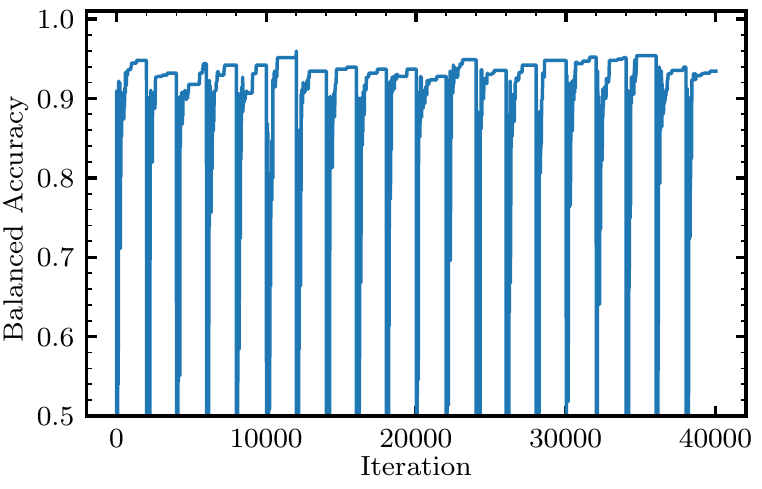}	
        \caption{Evolution of the objective function.}\label{fig:native_evolution_breast_cancer}
	\end{subfigure}
	\begin{subfigure}[t]{0.5\textwidth}
		\centering
		\includegraphics{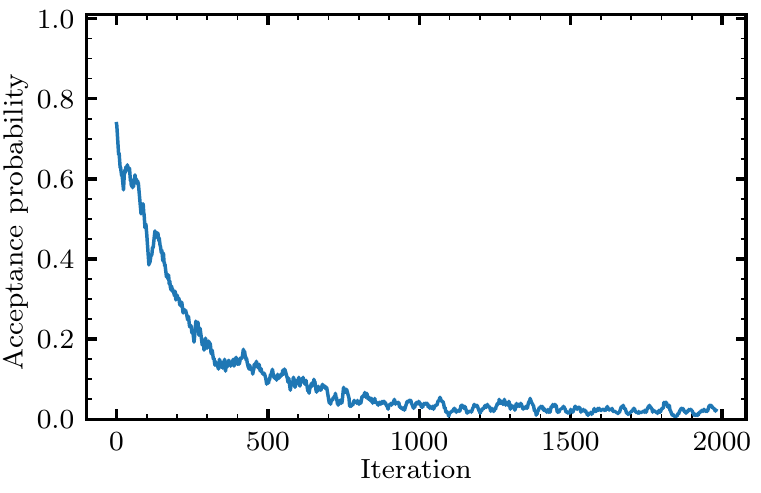}
        \caption{Evolution of the acceptance probability.}\label{fig:native_acceptance_probabilites_breast_cancer}
	\end{subfigure}
  \end{minipage}
  \caption{An example native local classifier run on the Breast Cancer dataset \cite{wolberg1992breast}. The settings for the solver are \codevar{num\_starts}~$=20$, \codevar{num\_iterations}~$=2000$, and the temperatures follow a geometric schedule from 0.2 to $10^{-6}$. The acceptance probability, which is the probability of accepting a proposed move, is  averaged across the starts and a window of length 20.}
\label{fig:example_solver_run}
\end{figure*}

The solver starts from a new random rule \codevar{num\_starts} times, which diversifies the search and is embarrassingly parallelizable. We have implemented a simulated annealing \cite{kirkpatrick1983optimization} solver, but other stochastic local search solvers \cite{hoos2004stochastic} could also be implemented (e.g., greedy, tabu, etc.). Each start begins by generating an initial random rule, and continues with the proposal of \codevar{num\_iterations} local moves. Moves are accepted based on a Metropolis criterion, such that the acceptance probability of a proposed move depends only on the objective function change and temperature. Initially, the temperature is high, leading to most moves being accepted (exploration). The temperature is decreased on a geometrical schedule such that, in the latter stages of each start, the solver accepts only descending or ``sideways'' moves---moves that do not change the objective function (exploitation). See \cref{fig:example_solver_run} for an example run. 

\subsection{Non-local moves}
\label{sec:nonlocal_moves}

The local native optimizer described above searches the formula space by making small \textit{local} moves. It is also possible to perform larger, \textit{non-local} moves. Such moves are more expensive computationally, but they have the potential to improve the objective function more drastically by extending the range of the otherwise local search, as well as allowing the solver to escape local minima. In this section, we describe ways of proposing good non-local moves that are compatible with, and inspired by, quantum algorithms for optimization that might show a speedup over classical algorithms, namely algorithms for solving QUBO and ILP problems \cite{durr1996quantum, farhi2014quantum, khosravi2021mixed, montanaro2020quantum}. 

The basic idea behind the proposed non-local moves is to make a move that optimizes an entire subtree of the current formula. Given a randomly-selected target node (an operator or literal) and a new operator, we assign the new operator to the target node and optimize the subtree beneath it. 

\begin{figure*}[htb]
  \begin{minipage}{\textwidth}
	\begin{subfigure}[t]{7cm}
		\centering
		\includegraphics[width=7cm]{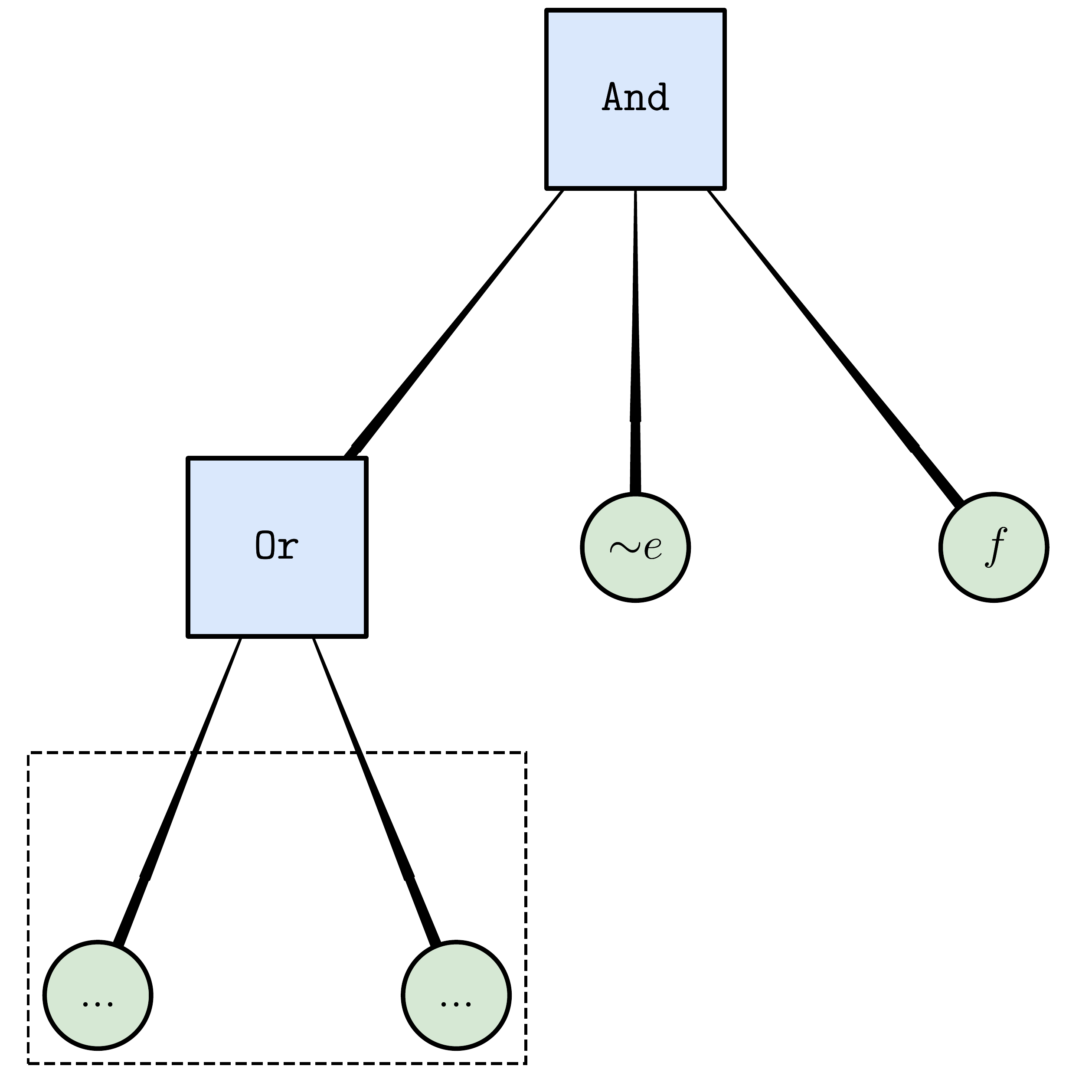}
		\caption{Swap operator \texttt{Choose2} for \texttt{Or} with a subtree of depth-one.}\label{fig:nonlocal_moves_a}
	\end{subfigure}\hspace{1cm}
	\begin{subfigure}[t]{7cm}
		\centering
		\includegraphics[width=7cm]{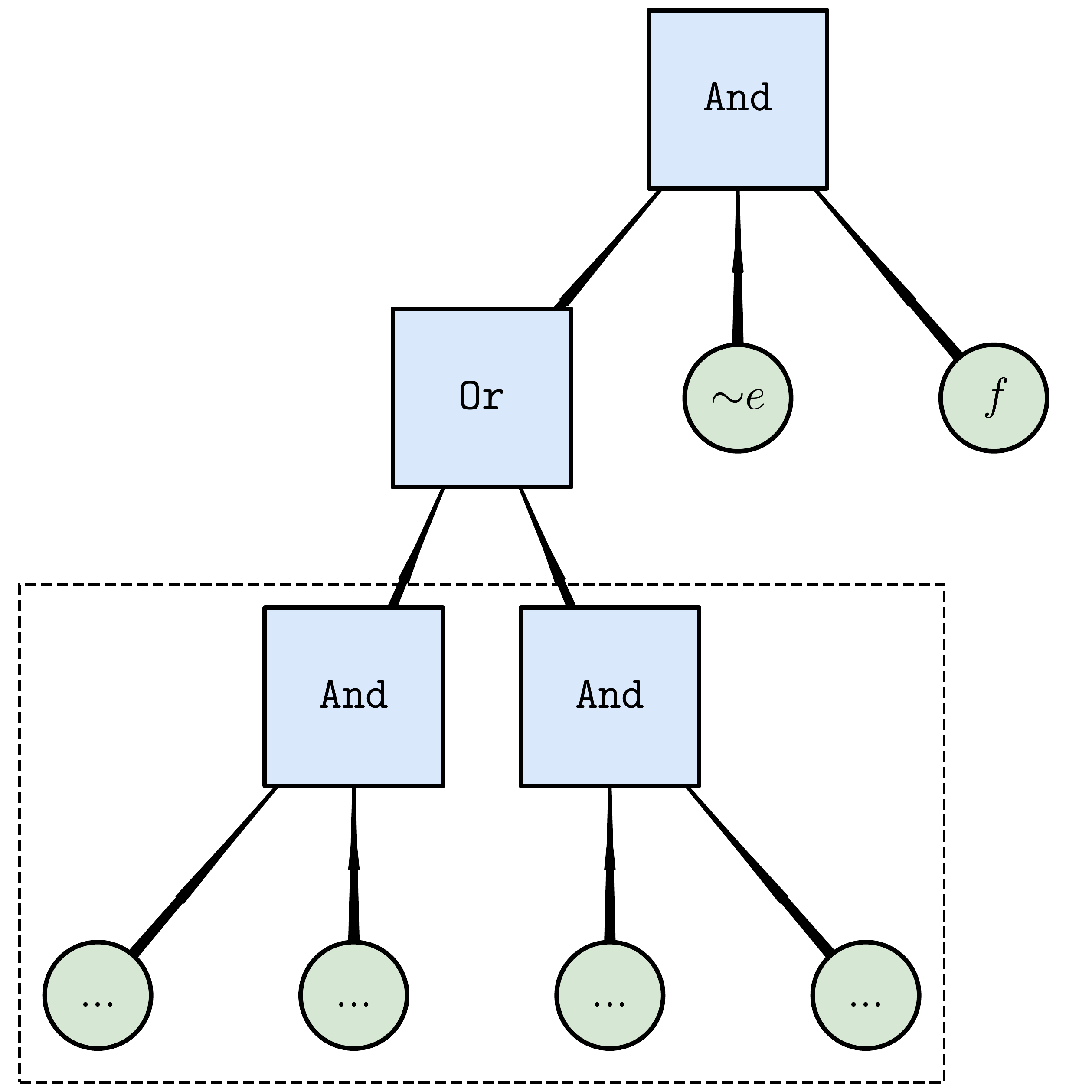}
		\caption{Swap operator \texttt{Choose2} for \texttt{Or} with a subtree of depth-two in DNF form.}\label{fig:nonlocal_moves_b}
	\end{subfigure}
  \end{minipage}
  \caption{Non-local moves---\textit{swap node with subtree of depth $d$}. These panels show examples of moves that replace an operator (or a literal, not shown) with a chosen operator (in this case \texttt{Or}) and an optimized depth-one (a) and depth-two (b) subtree. The latter shows an example disjunctive normal form move (\texttt{Or} of \texttt{And}s), but other structures are possible. Both moves are relative to the initial rule shown in \cref{fig:boolean_formula}. The dashed rectangle shows the subtree to be optimized. New literals are represented schematically, using dots; their actual number could vary.}
  \label{fig:nonlocal_moves}
\end{figure*}

The optimized subtree could be of depth-one (i.e., an operator with literals under it), or of depth-two (i.e., an operator with other operators and literals under it, with literals under those secondary operators). The optimization of the subtree could be limited to a particular subset of the possible subtrees, which can occur when using fixed format optimizers. An example of a depth-two tree of a fixed format is a DNF rule, i.e., an \texttt{Or} of \texttt{And}s. We refer to the subtree optimization move as \textit{swap node with subtree of depth $d$} [see \cref{fig:nonlocal_moves}(a)--(b)], and in this work we focus on the $d=1$ case.

In order to perform the optimization over the subtree $T$, we need to determine the effective input data and labels for the subtree optimization problem, $\bm{X'}$ and $\bm{y'}$, respectively. We now use the notation $R(T(x_i), x_i)$ in order to highlight that the evaluation of the rule can be thought of as being the result of evaluating the subtree, and then evaluating the rest of the rule given the result of the subtree evaluation. For each data row (sample) $x_i$ we evaluate the rule $R(T(x_i), x_i)$ twice, by first substituting $T(x_i)=0$, and then $T(x_i)=1$. If the result is the same in both cases, then it is \textit{predetermined}, and hence the sample $x_i$ does not contribute to the score of \textit{any} proposed subtree $T$. If the result is different, then we set $y_i'$ to the value that causes the classification to be correct, i.e., such that $R(x_i)=y_i$, and add the corresponding sample to $\bm{X'}$. Note that it is guaranteed in this case that one of these options classifies the sample correctly, because in this case we have two different outputs, and those cover all of the possible outputs (for binary classification). The subtree score is then calculated over only the labels that are not predetermined, i.e., the effective labels. The complexity of the subtree $C(T)$ is calculated in the same way as it is calculated for the complete rule. 

To determine the non-local move, we solve an optimization problem that is determined by fixing $R / T_0$ (i.e., everything that is not the target node or a descendant of it) in \cref{eq:rule_objective_function} and discarding constants, where $T_0$ is the target subtree (prior to optimization). In particular, we have
\begin{alignat}{2}
\label{eq:subtree_objective_function}
&T^{*} = \argmax_T[S(T(\bm{X'}), \bm{y'}) - \lambda \, C(T)] \\
&\text{s.t.} \quad C(T) \leq C' - [C(R) - C(T_0)], \notag
\end{alignat}
where $\bm{X'}$ and $\bm{y'}$ are the input data and effective subtree labels for the non-predetermined data rows, respectively, $T$ is any valid subtree, and $T^{*}$ is the optimized subtree (i.e., the proposed non-local move). 

In practice, we must also constrain the complexity of the subtree from below, because otherwise the optimized subtree could cause the rule to be invalid. For this reason, if the target is the root, we enforce ${\codevar{min\_num\_literals}=2}$, because the root operator must have two literals or more. If the target is not the root, we enforce ${\codevar{min\_num\_literals}=1}$, to allow the replacement of the subtree with a single literal (if beneficial). The ILP and QUBO formulations we present in \cref{sec:ilp_qubo_formulations} do not include this lower bound on the number of literals for simplicity, but its addition is trivial. 

Yet another practical consideration is that we 
we want to propose non-local moves fast, to keep the total solving time within the allotted limit. One way of trying to achieve this is to set a short timeout. However, the time required to construct and solve the problems is dependent on the number of non-determined samples. If the number of samples is large and the timeout is short, the best solution found might be of poor quality. With this in mind, we define a parameter named \codevar{max\_samples} which controls the maximum number of samples that can be included in this optimization problem. This parameter controls the tradeoff between the effort required to find a good non-local move to propose, and the speed with which such a move can be constructed. Even with this parameter in place, the time required to find the optimal solution can be significant (for example minutes or more), so we utilize a short timeout.

\subsection{Putting it all together}
\label{sec:nonlocal_moves_our_implementation}

\begin{figure}[htb]
\includegraphics[width=1.0 \columnwidth]{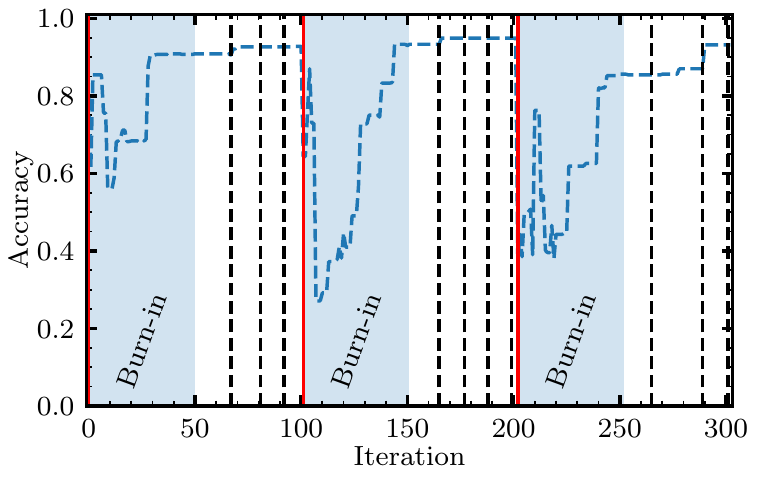}
\caption{Evolution of the objective function (in this case, the accuracy) in a short example run of the native local solver with non-local moves on the Breast Cancer dataset with \codevar{max\_complexity}~$=10$. The settings for the solver are \codevar{num\_starts}~$=3$, \codevar{num\_iterations}~$=100$, and the temperatures follow a geometric schedule from $0.2$ to $10^{-4}$. The first iteration of each start is indicated by a vertical solid red line. The first \codevar{num\_iterations\_burn\_in}~$=50$ iterations of each start are defined as the burn-in period (shaded in blue), in which no non-local moves are proposed. After that, non-local moves are proposed when there is no improvement to the accuracy in \codevar{patience}~$=10$ iterations. The proposals of non-local moves use a subset of the samples of size \codevar{max\_samples}~$=100$, and are indicated by vertical dashed black lines.}
\label{fig:example_native_non_local}
\end{figure}

Now that we have described the idea of the non-local moves at a high-level, we describe the way in which we have incorporated them into the native local solver in more detail. \cref{algo:solver} presents pseudo-code for the simulated annealing native local solver with non-local moves. Non-local moves are relatively expensive to calculate, and thus should be used sparingly. Moreover, non-local moves are fairly exploitative (as opposed to exploratory), so in the context of simulated annealing, proposing such moves at the beginning of the optimization would not help convergence, and therefore would be a waste of computational resources. With these points in mind, we define a burn-in period \codevar{num\_iterations\_burn\_in} for each start, in which no non-local moves are proposed. After the burn-in period, non-local moves are proposed only if a certain number of iterations (referred to as the \codevar{patience}) has gone by with no improvements due to proposing local moves. A short example run is presented in \cref{fig:example_native_non_local}, showing the evolution of the objective function over multiple starts, the way in which the non-local moves kick in after an initial burn-in period, and that they are indeed able to improve the objective function measurably (but also sometimes fail to do so). 

\begin{figure*}
\begin{center}
\noindent\begin{minipage}{.7\linewidth}
\lstset{caption={The pseudo-code for our native local solver with non-local moves. The solver executes \codevar{num\_starts} starts, each with \codevar{num\_iterations} iterations. In each start, a random initial rule is constructed, and then a series of local (see \cref{algo:propose_local_move}) and non-local moves (see \cref{sec:nonlocal_moves}) are proposed and accepted based on the Metropolis criterion. Non-local moves are only proposed after an initial \codevar{num\_iterations\_burn\_in} iterations and if there has been no improvement over \codevar{patience} iterations. The initial rule and the proposed moves are constructed such that the current rule is always feasible, and in particular has a complexity no higher than \codevar{max\_complexity}. Non-local moves replace an existing literal or operator with a subtree, optimized over a randomly selected subset of the data of size \codevar{max\_samples}. The solver returns the best rule found. Some details omitted due to lack of space.}}
\begin{lstlisting}[language=Python, label={algo:solver}]
def solve(X, y, max_complexity, num_starts, 
          num_iterations, num_iterations_burn_in, patience):
    best_score = -inf
    best_rule = None
    for start in range(num_starts):
        is_patience_exceeded = False
        current_rule = generate_initial_rule(X, max_complexity)
        current_score = score(current_rule, X, y)
        
        for iteration in range(num_iterations):
            T = update_temperature()
            if iteration > num_iterations_burn_in and is_patience_exceeded:
                proposed_move = propose_non_local_move(current_rule, max_samples)
            else:
                proposed_move = propose_local_move(current_rule)
            
            proposed_move_score = score(proposed_move, X, y)
            dE = proposed_move_score - current_score
            accept = dE >= 0 or random.random() < exp(dE / T)

            if accept:
                current_score = proposed_move_score
                current_rule = proposed_move
            
                if current_score > best_score:
                    best_score = current_score
                    best_rule = deepcopy(current_rule)

            is_patience_exceeded = update_patience_exceeded(patience)
            
    return best_rule
\end{lstlisting}
\end{minipage}
\end{center}
\end{figure*}

\section{Depth-one ILP and QUBO formulations}
\label{sec:ilp_qubo_formulations}

In this section we formulate the problem of finding optimal depth-one rules as an ILP or QUBO problem, with various operators at the root. We use these formulations in two ways: 1) to find optimized depth-one rules that form the basis of standalone classifiers, and 2) to find good non-local moves in the context of a native local solver, which periodically proposes non-local moves.  

We start by following Ref.~\cite{malioutov2017learning}, which explains how to formulate the search for optimal \texttt{Or}, \texttt{And}, and \texttt{AtLeast} rules as ILP problems. Our contributions here are: 
\begin{enumerate}
    \item The addition of negated features.
    \item Design of even-handed formulations (unbiased towards positive/negative samples), and the addition of class weights.
    \item Correction of the original formulation for \texttt{AtLeast}, and generalizations to the \texttt{AtMost} and \texttt{Choose} operators.
    \item Direct control over the score/complexity tradeoff by constraining the number of literals.
\end{enumerate}
We then explain how to translate these formulations into QUBO formulations. Finally, we describe how the constraints can be softened, thereby reducing the search space significantly.

\subsection{Formulating the \texttt{Or} rule as an ILP}
\label{sec:ilp_or}

We start by observing that the rule $y$ = \texttt{Or($f_0, f_1$)} can be expressed equivalently as:
\begin{alignat}{2}
\label{eq:ilp_idea_perfect_or}
&f_0 + f_1 \geq 1 & \quad \text{for } y=1\phantom{.} \\
&f_0 + f_1 = 0      & \quad \text{for } y=0. \notag
\end{alignat}
We can then define an optimization problem to find the smallest subset of features $f_i$ to include in the \texttt{Or} rule, in order to achieve perfect accuracy:
\begin{alignat}{2}
\label{eq:ilp_formulation_perfect_or}
&\!\min      \quad && ||\tb||_0 \\
&\text{s.t.} \quad && \tXp \tb \geq \bm{1} \notag \\
&                  && \tXn \tb =      \bm{0}, \notag \\
&                  && \tb \in \{0, 1\}^m \notag 
\end{alignat}
where $\tb$ is a vector of indicator variables, indicating for each feature whether it should be included in the rule (i.e., $b_{i}=1$ if feature $f_{i}$ is included and $b_{i}=0$ otherwise), $\tXp$ is a matrix containing only the rows labelled as ``positive'' ($y=1$), $\tXn$ is a matrix containing only the rows labelled as ``negative'' ($y=0$), and $\bm{0}$ and $\bm{1}$ are vectors containing only zeros and ones, respectively. 

We extend this formulation by adding the possibility to include negated features in the rule. While this could be accomplished simply by adding the negated features to the input data $\tX$, thereby doubling the size of this matrix, it may be more efficient to include the negation in the formulation. To accomplish this, we add an additional vector of indicator variables $\ttb$, indicating for each negated feature whether it should be included in the rule. We then replace $\tX \tb \to \tX \tb + \ttX \ttb$, noting that $\ttX = \bm{1} - \tX$ (where $\bm{1}$ is now a matrix of ones) since a binary variable $v\in\{0,1\}$ is negated by $1-v$. Therefore, we have:
\[
\tX \tb \to \tX \tb + \ttX \ttb = \tX \tb + (\bm{1} - \tX) \ttb = \tX (\tb - \ttb) +  ||\ttb||_0,
\]
where $||\ttb||_0$ is the sum over the entries of $\ttb$. By substituting the above into \cref{eq:ilp_formulation_perfect_or} and adding a corresponding term for $\ttb$ to the objective function, we find:
\begin{alignat}{2}
\label{eq:ilp_formulation_perfect_or_with_negation}
&\!\min      \quad && ||\tb||_0 + ||\ttb||_0 \\
&\text{s.t.} \quad &&  \tXp (\tb-\ttb) + ||\ttb||_0 \geq \bm{1} \notag \\
&                         &&  \tXn (\tb-\ttb) + ||\ttb||_0 = \bm{0} \notag \\
&                  && \tb, \ttb \in \{0, 1\}^m. \notag 
\end{alignat}

In practice, we typically do not expect to be able to achieve perfect accuracy. With this in mind, we introduce a vector of ``error'' indicator variables $\bm{\eta}$, indicating for each data row whether it is misclassified. When the error variable corresponding to a particular sample is 1, the corresponding constraint is always true by construction, effectively deactivating that constraint.  Accordingly, we change our objective function such that it minimizes the number of errors. To control the complexity of the rule, we add a regularization term, as well as an explicit constraint on the number of literals. Finally, in order to deal with unbalanced datasets, we allow the positive and negative error terms to be weighted differently:
\begin{alignat}{2}
\label{eq:ilp_formulation_imperfect_or_with_negation_with_max_num_literals}
&\!\min      \quad && (\twp||\bm{\tetap}||_0 + \twn||\bm{\tetan}||_0) + \lambda (||\tb||_0 + ||\ttb||_0) \\
&\text{s.t.} \quad && \tXp (\tb-\ttb) + ||\ttb||_0 + \tetap \geq \bm{1} \notag \\
&                         && \tXn (\tb-\ttb) + ||\ttb||_0 - \tetan m'\leq \bm{0} \notag \\
&			  &&  ||\tb||_0 + ||\ttb||_0 \leq m' \notag \\
&             && \tb, \ttb \in \{0, 1\}^m, \bm{\eta} \in \{0, 1\}^n, \notag 
\end{alignat}
where $m'$ (also referred to as \codevar{max\_num\_literals}) is the maximum number of literals allowed, $m' \leq m$, where $m$ is the number of features, and $\twp$ and $\twn$ are the positive and negative class weights, respectively. The default choice for the class weights is to make them inversely proportional to the fraction of samples of each respective class. In this work, we set $\twp = n/(2\np)$ and similarly $\twn = n/(2\nn)$, where $\np$ is the number of positive samples, $\nn$ is the number of negative samples, and $n$ is the total number of samples ($n=\np + \nn$).

Inspecting \cref{eq:ilp_formulation_imperfect_or_with_negation_with_max_num_literals}, one can see that when a positive data row is misclassified ($\eta=1$), then the first constraint is always fulfilled. This is because the left hand side of that constraint is then the sum of one plus a non-negative number, which is always larger than or equal to one. Similarly, when a negative data row is misclassified, then the second constraint is always satisfied. This is because the left hand side of that constraint consists of a number that is bound from above by $m'$ from which $m'$ is subtracted, which is surely negative or zero. Notice that this second constraint, for the negative samples, was previously an equality constraint [in \cref{eq:ilp_formulation_perfect_or_with_negation}]. It was changed to an inequality constraint in order to make the ``deactivation'' (for $\eta=1$) work.  

\subsection{Formulating the \texttt{And} rule as an ILP}
\label{sec:ilp_and}

In the previous section we described the ILP formulation for the \texttt{Or} operator. In this section we show how this formulation can be extended to the \texttt{And} operator. It is instructive to first consider why it is useful to go beyond the \texttt{Or} rules at all, and what is the exact nature of the relationship between \texttt{Or} and \texttt{And} rules in this context. 

\begin{figure*}[htb]
  \begin{minipage}{\textwidth}
	\begin{subfigure}[t]{0.45\textwidth}
	        \centering
		\includegraphics[width=\textwidth]{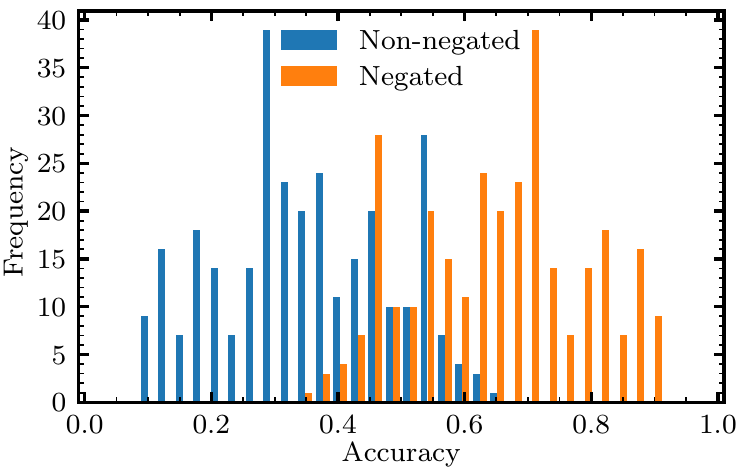}
		\caption{Single feature rules}\label{fig:score_landscape_single}
	\end{subfigure}\hspace{0.5cm}
	\begin{subfigure}[t]{0.47\textwidth}
		\centering
		\includegraphics[width=\textwidth]{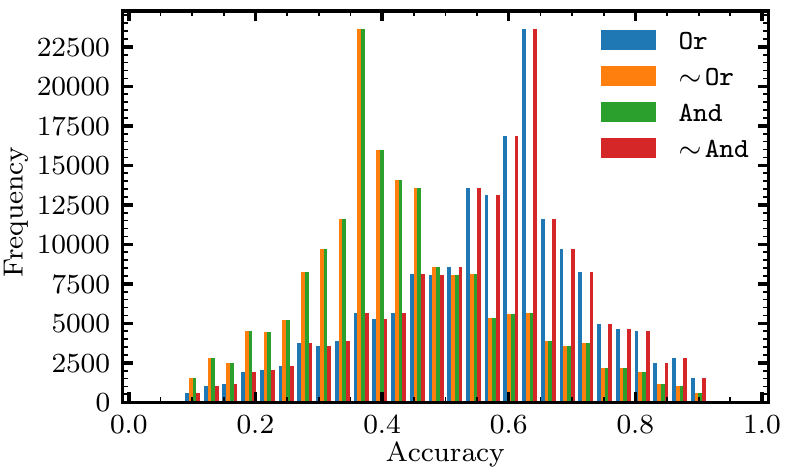}
		\caption{Double feature rules}\label{fig:score_landscape_double}
	\end{subfigure}
  \end{minipage}
  \caption{Accuracy score landscape for single and double feature rules for the full Breast Cancer dataset. Rules and respective scores are obtained by enumerating the full search space.}
  \label{fig:score_landscape}
\end{figure*}

We note that De Morgan's laws (which we shall use below) guarantee that every \texttt{Or} rule has an equivalent $\mathord\sim\texttt{And}$ rule (and similarly for \texttt{And} and $\mathord\sim\texttt{Or}$). The relation between the distribution of the different rules is apparent from \cref{fig:score_landscape} where we plot the score landscape for both single feature and double feature rules. In particular, the distribution of \texttt{Or} clearly matches the distribution of \msim\texttt{And} (and similarly for \texttt{And} and \msim \texttt{Or}), as expected. More importantly for us, the distribution of \texttt{Or} is clearly very different from the distribution of \texttt{And}, motivating the introduction of the latter. Similarly, the distribution of non-negated single feature rules is clearly very different from the distribution of negated single feature rules, again motivating the inclusion of negated features. As an aside, note the reflection symmetry in both plots, which is due to the identity $S(R) = 1-S(\mathord\sim R)$.

The best (optimal) rules for each distribution are given in \cref{tab:score_landscape_best_rules}. The best \texttt{And} rule is significantly better than the best \texttt{Or} rule in this case, clearly motivating its inclusion. Furthermore, the reason for including negated features is clear from comparing the best single feature rule scores. 

As a final motivational observation, we note that as \codevar{max\_num\_literals} is increased, \texttt{Or} rules generally tend to produce false positives (since additional features tend to push more outputs to one), whereas \texttt{And} rules generally tend to produce false negatives (since additional features tend to push more outputs to zero). Different use cases might lend themselves to either of these operators. However, both of these operators are both fairly rigid, a point which is mitigated by the introduction of parametrized operators (see \cref{sec:ilp_parameterized_operators}), and the introduction of higher depth rules, as found by the native local solver (see \cref{sec:native_optimization}). 

\begin{table*}[htb]
	\centering
 	\caption{The best (optimal) rules for single and double feature rules for the full Breast Cancer dataset. ``Type'' is the type of rule, ``Rules'' is the number of rules of that rule type, ``Accuracy'' is regular accuracy (the metric used for this table), and ``Rule'' is one of the optimal rules for that respective rule type. Negations of features are post-processed out for readability by reversing the relationship in the respective feature name (e.g., $\mathord\sim a>3 \to a \leq 3$).\label{tab:score_landscape_best_rules}}
	\begin{tabular}{l@{\hskip 0.1in}r@{\hskip 0.1in}c@{\hskip 0.1in}l}
		\toprule
		Type & Rules & Accuracy & Rule \\
		\midrule
		\phantom{\msim}$f$               & 300     & 0.647 & mean fractal dimension~$>$~0.0552 \\
		\msim$f$                         & 300     & 0.914 & worst perimeter~$\leq$~108.9364 \\
  \midrule
		\phantom{\msim}\texttt{Or}       & 179,400 & 0.931 & \texttt{Or(}worst concave points~$\leq$~0.1091, worst area~$\leq$~719.6364\texttt{)} \\
        \msim\texttt{Or}                 & 179,400 & 0.944 & \msim \texttt{Or(}worst concave points~$>$~0.1563, worst area~$>$~988.6818\texttt{)} \\
        \phantom{\msim}\texttt{And}      & 179,400 & 0.944 & \texttt{And(}worst concave points~$\leq$~0.1563, worst area~$\leq$~88.6818\texttt{)} \\
        \msim\texttt{And}                & 179,400 & 0.931 & \msim \texttt{And(}worst concave points~$>$~0.1091, worst area~$>$~719.6364\texttt{)} \\
		\bottomrule
	\end{tabular}
\end{table*}

To formulate the \texttt{And} operator, we use De Morgan's laws. Namely, starting from \cref{eq:ilp_formulation_imperfect_or_with_negation_with_max_num_literals} we swap $\tb \leftrightarrow \ttb$, $\tetan \leftrightarrow \tetap$, and $\tXn \leftrightarrow \tXp$, to find:
\begin{alignat}{2}
\label{eq:ilp_formulation_imperfect_and_with_negation_with_max_num_literals}
&\!\min      \quad && (\twp||\bm{\tetap}||_0 + \twn||\bm{\tetan}||_0) + \lambda (||\tb||_0 + ||\ttb||_0) \\
&\text{s.t.} \quad && \tXn (\ttb-\tb) + ||\tb||_0 + \tetan \geq \bm{1} \notag \\
&                         && \tXp (\ttb-\tb) + ||\tb||_0 - \tetap m'\leq \bm{0} \notag \\
&			  &&  ||\tb||_0 + ||\ttb||_0 \leq m' \notag \\
&             && \tb, \ttb \in \{0, 1\}^m, \bm{\eta} \in \{0, 1\}^n. \notag 
\end{alignat}

\subsection{Extending the ILP formulation to parametrized operators}
\label{sec:ilp_parameterized_operators}

So far we have shown how the problem of finding an optimal depth-one rule with \texttt{Or} and \texttt{And} operators in the root can be formulated as an ILP. Here we extend these formulations to the parameterized operators \texttt{AtLeast}, \texttt{AtMost}, and \texttt{Choose}. This is motivated by the hypothesis that the additional parameter in these operators could provide additional expressivity, which could translate into better results for particular datasets, at least for some values of \codevar{max\_num\_literals}. 

\begin{figure*}[htb]
\includegraphics[width=1.0\linewidth]{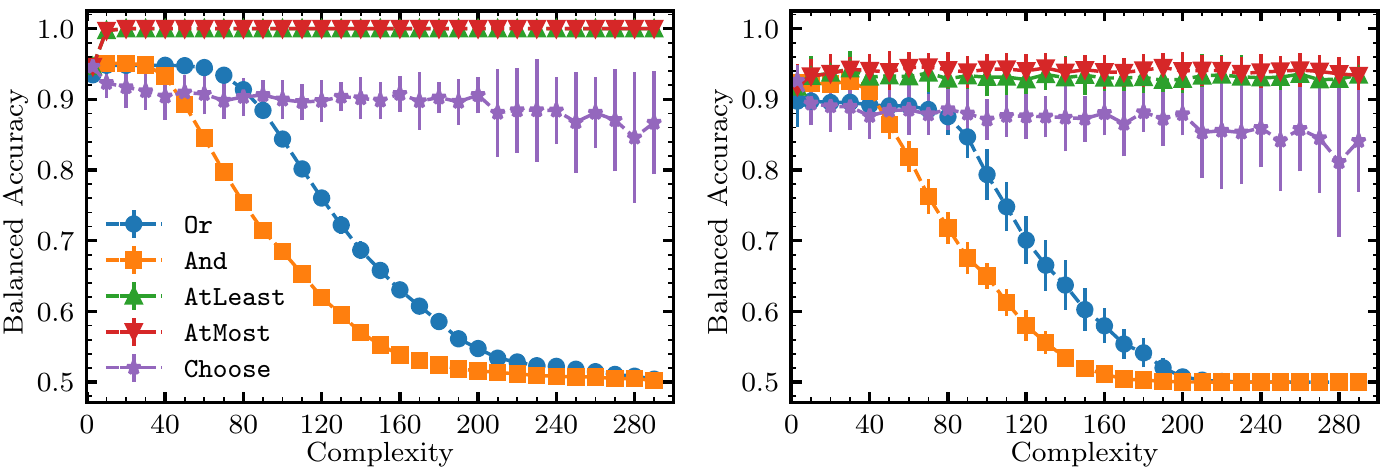}
\caption{A comparison of the train (left) and test (right) expressivity of different operators in depth-one rules, solved via the ILP formulations. Each classifier is trained and tested on 32 stratified shuffled splits with a 70/30 split (for cross-validation) over the in-sample data (80/20 stratified split). The points correspond to the mean of the balanced accuracy and complexity over those splits. The error bars are given by the standard deviation over the balanced accuracy and complexity on the respective 32 splits for each point. }
\label{fig:ilp_full_range_results_breast_cancer}
\end{figure*}

We probe this idea in an experiment, in which we vary \codevar{max\_num\_literals} over the full possible range for the Breast Cancer dataset  in \cref{fig:ilp_full_range_results_breast_cancer}. In this experiment, we fix \codevar{max\_num\_literals}=\codevar{min\_num\_literals} to expose the true objective value over the full range. One can clearly see that the additional expressivity of the parameterized operators allows them to maintain a higher accuracy over most of the possible range, for both the train and test samples. For the unparameterized operators, we observe significantly reduced performance at high \codevar{max\_num\_literals}, explained by the fact that eventually the solver runs out of features that can be added and still improve the score. For the parameterized operators, this limitation is mitigated by adjusting the corresponding parameter. 

To formulate the \texttt{AtLeast} operator, we modify \cref{eq:ilp_idea_perfect_or} such that the rule $y$ = \texttt{AtLeastk($f_0, f_1$)} can be expressed equivalently as
\begin{alignat}{2}
\label{eq:ilp_idea_perfect_atleast}
&f_0 + f_1 \geq k    & \quad \text{for } y=1\phantom{.} \\
&f_0 + f_1 \leq k-1  & \quad \text{for } y=0. \notag
\end{alignat}
Accordingly, we modify \cref{eq:ilp_formulation_imperfect_or_with_negation_with_max_num_literals} to obtain the ILP formulation for \texttt{AtLeast}:
\begin{alignat}{2}
\label{eq:ilp_formulation_imperfect_atleast_with_negation_with_max_num_literals}
&\!\min      \quad && (\twp||\bm{\tetap}||_0 + \twn||\bm{\tetan}||_0) + \lambda (||\tb||_0 + ||\ttb||_0) \\
&\text{s.t.} \quad && \tXp (\tb-\ttb) + ||\ttb||_0 + \tetap m' \geq k \bm{1} \notag \\
&                         && \tXn (\tb-\ttb) + ||\ttb||_0 - \tetan (m' + 1) \leq (k-1)\bm{1} \notag \\
&			  &&  0 \leq k \leq ||\tb||_0 + ||\ttb||_0 \leq m' \notag \\
&             && \tb, \ttb \in \{0, 1\}^m, \bm{\eta} \in \{0, 1\}^n, k \in \mathcal{Z} \notag 
\end{alignat}
noting that $k$ is a decision variable that is optimized over by the solver, rather than being chosen in advance. 

Similarly, we can formulate \texttt{AtMost} as:
\begin{alignat}{2}
\label{eq:ilp_formulation_imperfect_atmost_with_negation_with_max_num_literals}
&\!\min      \quad && (\twp||\bm{\tetap}||_0 + \twn||\bm{\tetan}||_0) + \lambda (||\tb||_0 + ||\ttb||_0) \\
&\text{s.t.} \quad && \tXp (\tb-\ttb) + ||\ttb||_0 - \tetap m' \leq k \bm{1} \notag \\
&                         && \tXn (\tb-\ttb) + ||\ttb||_0 + \tetan (m'+1) \geq (k+1)\bm{1} \notag \\
&			  &&  0 \leq k \leq ||\tb||_0 + ||\ttb||_0 \leq m' \notag \\
&             && \tb, \ttb \in \{0, 1\}^m, \bm{\eta} \in \{0, 1\}^n, k \in \mathcal{Z} \notag 
\end{alignat}
Note that any \texttt{AtLeastk} rule has an equal-complexity equivalent \texttt{AtMost} rule that can be obtained by taking $k \to F-k$ and $f_i\to \mathord\sim f_i$ for each feature $f_i$ in the original rule, where $F$ is the number of features in the original rule. For example, {\texttt{AtLeastk($f_0, f_1$)}} is equivalent to {\texttt{AtMost[2-k]($\mathord\sim f_0, \mathord\sim f_1$)}}. For this reason, we expect similar numerical results for \texttt{AtLeast} and \texttt{AtMost}. However, the actual rules differ, and a user might prefer one over the other, for example due to a difference in effective interpretability. 

Finally, formulating \texttt{Choose} is more complicated, because we need to formulate a not-equal constraint. Let us first write the equivalent of the perfect \texttt{Or} classifier of \cref{eq:ilp_formulation_perfect_or_with_negation} (not in ILP form yet):
\begin{alignat}{2}
\label{eq:ilp_formulation_perfect_choose_with_negation_neq}
&\!\min      \quad && (\twp||\bm{\tetap}||_0 + \twn||\bm{\tetan}||_0) + \lambda (||\tb||_0 + ||\ttb||_0) \\
&\text{s.t.} \quad &&  \tXp (\tb-\ttb) + ||\ttb||_0 = k \bm{1} \notag \\
&                         &&  \tXn (\tb-\ttb) + ||\ttb||_0 \neq k \bm{1} \notag \\
&			  &&  0 \leq k \leq ||\tb||_0 + ||\ttb||_0 \leq m' 
\notag \\
&             && \tb, \ttb \in \{0, 1\}^m, \bm{\eta} \in \{0, 1\}^n, k \in \mathcal{Z} \notag 
\end{alignat}
For the first constraint, we note that any equality constraint $a=b$ can be split equivalently into two inequalities, $a \leq b$ and $a \geq b$. Then, we already know how to add error variables to inequalities, as we did for the other formulations. Similarly, we can split the not-equal constraint into two inequalities, because any not-equal constraint $a \neq b$ over integers can be split into two disjunctive inequalities, $a \geq b+1$ and $a \leq b-1$. Once again, we already know how to add error variables to inequality constraints. However, in this case the constraints are mutually exclusive, and adding both of them causes our model to be infeasible. There is a well-known trick for modelling either-or constraints that can be applied here \cite{aimms2016aimms}. We add a vector of indicator variables $\bm{q}$ that chooses which constraint of the two to apply, for each negative sample. We end up with:
\begin{alignat}{2}
\label{eq:ilp_formulation_imperfect_choose_with_negation_with_max_num_literals}
&\!\min      \quad (\twp||\bm{\tetap}||_0 + \twn||\bm{\tetan}||_0) + \lambda (||\tb||_0 + ||\ttb||_0) \notag \\
&\text{s.t.} \notag \\
& \tXp (\tb-\ttb) + ||\ttb||_0 + \tetap m' \geq k\bm{1} \\
& \tXp (\tb-\ttb) + ||\ttb||_0 - \tetap m' \leq k\bm{1} \notag \\
& \tXn (\tb-\ttb) + ||\ttb||_0 + (\tetan + \bm{q}) (m'+1) \geq (k+1)\bm{1} \notag \\
& \tXn (\tb-\ttb) + ||\ttb||_0 - (\tetan + \bm{1}-\bm{q}) (m'+1) \leq (k-1)\bm{1} \notag \\
& 0 \leq k \leq ||\tb||_0 + ||\ttb||_0 \leq m' \notag \\
& \tb, \ttb \in \{0, 1\}^m, \bm{\eta} \in \{0, 1\}^n, \bm{q} \in \{0, 1\}^{\nn}, k \in \mathcal{Z} \notag 
\end{alignat}
One can think of \texttt{Choose} as being equivalent to a combination of \texttt{AtLeast} and \texttt{AtMost}. Accordingly, one can readily see that \cref{eq:ilp_formulation_imperfect_choose_with_negation_with_max_num_literals} is equivalent to a combination of \cref{eq:ilp_formulation_imperfect_atleast_with_negation_with_max_num_literals} (if $\bm{q}=\bm{0}$) and \cref{eq:ilp_formulation_imperfect_atmost_with_negation_with_max_num_literals} (if $\bm{q}=\bm{1}$).

\subsection{Converting the ILPs to QUBO problems}
\label{sec:qubo_formulations}

We seek to convert the above ILPs to corresponding QUBO problems, motivated by the fact that many quantum algorithms and devices are tailored towards QUBO problems \cite{farhi2014quantum, hauke2020perspectives, temme2011quantum, baritompa2005grover, tilly2022variational}. We start by describing the standard method \cite{glover2022quantum}, argue for a subtle change, and then lay out clearly a general recipe for converting an ILP to a QUBO problem. 

A QUBO problem is commonly defined as
\begin{alignat}{2}
&\!\min       \quad && \bm{x}^T\bm{Q} \bm{x} \\
&\text{s.t.}        && \bm{x} \in \{0, 1\}^N \notag
\end{alignat}
where $\bm{x}$ is a vector of binary variables, and $\bm{Q}$ is a real matrix \cite{glover2022quantum}. The standard method of including equality constraints such as $\bm{a}^T \bm{x}=b$ in a QUBO is to add a squared penalty term $P(\bm{a}^T \bm{x}-b)^2$, where $\bm{a}$ is a real vector, $b$ is a real number, and $P$ is a positive penalty coefficient. 

We now describe how to include inequality constraints such as $\bm{a}^T\bm{x} \leq b$ (without loss of generality) in a QUBO. We first note that $\bm{a}^T\bm{x}$ is bound from above by the sum of positive entries in $\bm{a}$ (denoted by $a_{+}$) and from below by the sum of negative entries in $\bm{a}$ (denote by $a_{-}$.) For the constraint $\bm{a}^T\bm{x} \leq b$, the assumption is that $b < a_{+}$, i.e., that $b$ provides a tighter bound (or else this constraint is superfluous). Therefore, we can write any inequality constraint in the form $l \leq \bm{a}^T\bm{x} \leq u$, where $l$ and $u$ are the lower and upper bounds respectively, which is the form we use for the rest of this section. 

We can convert this inequality constraint into an equality constraint by introducing an integer slack variable $0 \leq s \leq u-l$, which can be represented via a binary encoding. We can then use the above squaring trick to find the corresponding penalty term. It is important to note that there are two ways to do so:
\begin{alignat}{2}
&\bm{a}^T\bm{x}= l + s & \quad \text{(from below)}\phantom{.} \\
&\bm{a}^T\bm{x} = u - s & \quad \text{(from above)}. \notag
\end{alignat}
These equality constraints could then be included by adding the respective penalty term
\begin{alignat}{2}
&P(\bm{a}^T\bm{x} - l - s)^2 & \quad \text{(from below)}\phantom{.} \\
&P(\bm{a}^T\bm{x} - u + s)^2 & \quad \text{(from above)}. \notag
\end{alignat}
This brings us to a subtle point that is not commonly discussed. For an exact solver, it should not matter, in principle, which of the two forms of the penalty terms we choose to add. However, when using many heuristic solvers, it is desirable to reduce the magnitude of the coefficients in the problem. In the case of quantum annealers, the available coefficient range is limited, and larger coefficients require a larger scaling factor to reduce the coefficients down to a fixed range \cite{yarkoni2022quantum}. The larger the scaling factor, the more likely it is that some scaled coefficients will be within the noise threshold \cite{dwave_ice}. In addition, for many Monte Carlo algorithms, such as simulated annealing, larger coefficients require higher temperatures to overcome, which could lead to inefficiencies. 

In order to reduce the size of coefficients, we note that the above formulations are almost the same; they differ only in the sign in front of the slack variable $s$, which does not matter for this discussion, and the inclusion of either $l$ or $u$ in the equation. This motivates us to recommend choosing the penalty term that contains the bound ($l$ or $u$) that has a smaller absolute magnitude, because this yields smaller coefficients when the square is expanded. 

Based on the above, we now provide a compact recipe for converting ILP problems to QUBO problems:
\begin{enumerate}

\item Assume we have a problem in canonical ILP form:
\begin{alignat}{2}
&\!\max      \quad && \bm{c}^T \bm{x} \\
&\text{s.t.} \quad && \bm{A}\bm{x} \leq \bm{b} \notag \\
&                  && \bm{x} \geq \bm{0}, \bm{x} \in \mathcal{Z}^N. \notag
\end{alignat}

\item Convert the inequality constraints to equivalent equality constraints by the addition of slack variables:
\begin{alignat}{2}
&\!\max      \quad && \bm{c}^T \bm{x} \\
&\text{s.t.} \quad && \bm{A}\bm{x} = \bm{b} - \bm{s} \notag \\
&                  && \bm{x} \geq \bm{0}, \bm{x} \in \mathcal{Z}^N, \notag
\end{alignat}
where we have adopted, without loss of generality, the ``from above'' formulation.

\item Convert to a QUBO:
\begin{alignat}{2}
\label{eq:qubo_recipe}
&\!\min       \quad && \bm{x}^T(\bm{Q_{\mathrm{cost}} + Q_{\mathrm{penalty}}})\bm{x} \\
&\text{s.t.}        && \bm{x} \in \{0, 1\}^N \notag \\ 
&\text{where} \quad && \bm{Q_{\mathrm{cost}}} = -\diag(\bm{c}) \notag \\
&                   && \bm{Q_{\mathrm{penalty}}} = P\left\{\bm{A}^T \bm{A} - 2 \cdot \diag[\bm{A}^T (\bm{b}-\bm{s})]\right\}, 
\notag
\end{alignat}
where we have dropped a constant, and $\diag(\bm{c})$ is a square matrix with $\bm{c}$ on the diagonal.
\end{enumerate}

Using this recipe, it is possible to translate each of the five ILP formulations to corresponding QUBO formulations. As a representative example, we show how to do so for the \texttt{Or} formulation. We start from \cref{eq:ilp_formulation_imperfect_or_with_negation_with_max_num_literals}, which is reproduced here for easier reference:
\begin{alignat}{2}
\label{eq:ilp_formulation_imperfect_or_with_negation_with_max_num_literals2}
&\!\min      \quad && (\twp||\bm{\tetap}||_0 + \twn||\bm{\tetan}||_0) + \lambda (||\tb||_0 + ||\ttb||_0) \\
&\text{s.t.} \quad && \tXp (\tb-\ttb) + ||\ttb||_0 + \tetap \geq \bm{1} \notag \\
&                         && \tXn (\tb-\ttb) + ||\ttb||_0 - \tetan m'\leq \bm{0} \notag \\
&			  &&  ||\tb||_0 + ||\ttb||_0 \leq m' \notag \\
&             && \tb, \ttb \in \{0, 1\}^m, \bm{\eta} \in \{0, 1\}^n. \notag 
\end{alignat}
Upon applying the conversion recipe, we find the following QUBO:
\begin{alignat}{2}
\label{eq:qubo_formulation_or}
&\!\min      \quad && (\twp||\bm{\tetap}||_0 + \twn||\bm{\tetan}||_0) + \lambda (||\tb||_0 + ||\ttb||_0) + \\
&             && \twp L_1 \left\{ \tXp (\tb-\ttb) + ||\ttb||_0 + \tetap - \bm{1} - \bm{s} \right\}^2 + \notag \\
&             && \twn L_1 \left\{ \tXn (\tb-\ttb) + ||\ttb||_0 - \tetan m' + \bm{r} \right\}^2 + \notag \\
&			  &&  L_2 \left[||\tb||_0 + ||\ttb||_0 -m' - t\right]^2 \notag \\
&\text{s.t.} \quad && \tb, \ttb \in \{0, 1\}^m, \bm{\eta} \in \{0, 1\}^n, \notag \end{alignat}
where the curly brackets should be interpreted as a sum over the rows of the vector expression within the brackets, $\bm{s}$ and $\bm{r}$ are vectors of slack variables, $t$ is a slack variable, and $L_1$ and $L_2$ are positive penalty coefficients. The strength of the maximum number of literals constraint should be much larger than the strength of the soft constraints, to ensure it is enforced, i.e., $L_2 \gg L_1$. We do not write out the matrices in the QUBO formulation $\bm{Q_{\mathrm{cost}}}$ and $\bm{Q_{\mathrm{penalty}}}$ explicitly here, but they can readily be identified from \cref{eq:qubo_formulation_or}.

\subsection{Reducing the number of variables}

In this section we describe a way of reducing the number of variables in the QUBO formulation by eliminating the error variables. 

Recall that we introduced the misclassification indicator variables $\eta$ in order to soften the constraints. This inclusion made sense for the ILP formulation in which the constraints would otherwise be applied as hard constraints. In QUBO formulations, the difference between soft and hard constraints is just in the magnitude of the penalty coefficients. Therefore, we can, in principle, eliminate the error variables from the ``imperfect'' formulations and then add the constraints as soft constraints by construction, setting the penalty coefficients for those constraints to be relatively small. 

For example, for the \texttt{Or} operator, we take \cref{eq:ilp_formulation_imperfect_or_with_negation_with_max_num_literals2}, delete the error variables from the objective function and constraints, and label those constraints as ``soft'':
\begin{alignat}{2}
\label{eq:ilp_formulation_imperfect_or_with_negation_with_max_num_literals_no_eta_intermediate}
&\!\min      \quad && \lambda (||\tb||_0 + ||\ttb||_0) \\
&\text{s.t.} \quad &&  \tXp (\tb-\ttb) + ||\ttb||_0 \geq \bm{1} \quad \text{(Soft)} \notag \\
&                         &&  \tXn (\tb-\ttb) + ||\ttb||_0 =      \bm{0} \quad \text{(Soft)} \notag \\
&			  &&  ||\tb||_0 + ||\ttb||_0 \leq m' \notag \\
&             && \tb, \ttb \in \{0, 1\}^m. \notag 
\end{alignat}
We then apply the QUBO to ILP recipe described above to each constraint, and add the class weights to find:
\begin{alignat}{2}
\label{eq:ilp_formulation_imperfect_or_with_negation_with_max_num_literals_no_eta}
&\!\min      \quad && \lambda (||\tb||_0 + ||\ttb||_0) + \\
&             && \twp L_1 \left\{ \tXp (\tb-\ttb) + ||\ttb||_0 - \bm{1} - \bm{s} \right\}^2 + \notag \\
&             && \twn L_1 \left\{ \tXn (\tb-\ttb) + ||\ttb||_0 \right\}^2 + \notag \\
&			  &&  L_2 \left[||\tb||_0 + ||\ttb||_0 - m' - t\right]^2 \notag \\
&\text{s.t.} \quad && \tb, \ttb \in \{0, 1\}^m, \notag 
\end{alignat}
where we use the same curly bracket notation used in \cref{eq:qubo_formulation_or}, $\bm{s}$ is a vector of slack variables, $t$ is a slack variable, and $L_1$ and $L_2$ are positive penalty coefficients. As before, the strength of the maximum number of literals constraint should be much larger than the strength of the soft constraints, to ensure it is enforced, i.e., $L_2 \gg L_1$.

\begin{table}[htb]
	\centering
 	\caption{Number of variables required for various QUBO and ILP formulations. ``Operator'' is the operator at the root of the depth-one rule, ``with $\eta$'' refers to the QUBO formulation in which misclassifications are allowed via additional error variables, ``without $\eta$'' refers to the QUBO formulation in which misclassifications are allowed by soft constraints. The numbers quoted are for the complete Breast Cancer dataset which has 63\% negative labels, with \codevar{max\_num\_literals}~$=4$.\label{tab:num_variables_qubo_ilp}}
	\begin{tabular}{@{}l@{\hskip 0.2in}c@{\hskip 0.2in}c@{\hskip 0.2in}c@{}}
		\toprule
   		Operator        & with $\eta$ & without $\eta$ & ILP \\
		\midrule
		\texttt{Or}        & 2,879 & 1,027 & 1,169 \\ 
		\texttt{And}       & 2,879 & 1,317 & 1,169 \\ 
		\texttt{AtLeast}   & 3,097 & 1,959 & 1,170 \\ 
		\texttt{AtMost}    & 3,097 & 1,959 & 1,170 \\ 
		\texttt{Choose}    & 3,811 & 2,673 & 1,527 \\ 
		\bottomrule
	\end{tabular}
\end{table}

The reduction in problem size due to softening the constraints is generally significant (see \cref{tab:num_variables_qubo_ilp}). For example, for the \texttt{Or} formulation, we save the addition of the $\eta$ variables, one per sample. However, the dominant savings are in the slack variables for the negative data rows, because those constraints become equality constraints, thus avoiding the need for slack variables. In addition, there is a reduction in the range of the slack variables for the positive data rows, which can result in an additional reduction. The number of variables for each formulation is given by 
\begin{alignat}{2}
\label{eq:num_variables_or_qubo}
&\text{\texttt{Or} with} \,\, \eta\text{:} \\
&\text{\codevar{num\_vars}} = 2 m + n + \ceil{\log_2(m'+1)} (n+1) \notag \\
&\text{\texttt{Or}  without} \,\, \eta\text{:} \notag \\
&\text{\codevar{num\_vars}} = 2 m + \ceil{\log_2(m'+1)} + \ceil{\log_2(m')} \np. \notag
\end{alignat}
We hypothesize that the reduced problem size likely leads to a reduced time to solution (TTS). The TTS is commonly calculated as the product of the average time taken for a single start $\tau$, and the number of starts required to find an optimum with a certain confidence level, usually 99\%, referred to as $R_{99}$, i.e., ${\text{TTS}=\tau R_{99}}$ \cite{aramon2019physics}. The reduction in problem size should yield a shorter time per iteration, resulting in a smaller $\tau$. In addition, one might expect the problem difficulty to be reduced, resulting in a smaller $R_{99}$, but this is not guaranteed, as smaller problems are generally easier, but not always. 

\begin{figure*}[htb]
  		\centering
		\includegraphics[width=1.0\textwidth]{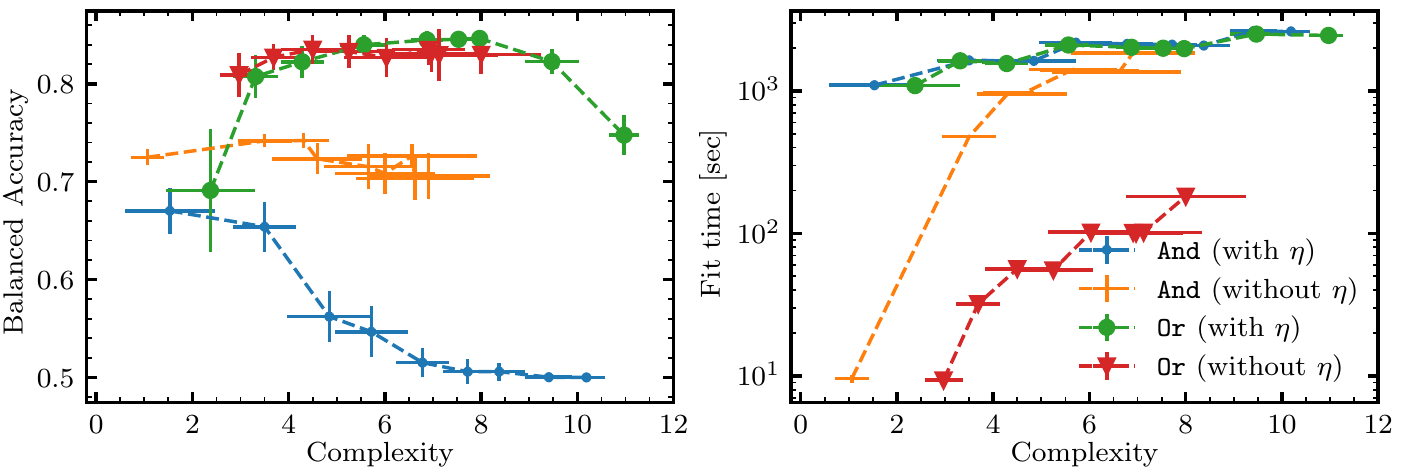}
  \caption{Test and timing results for the QUBO depth-one classifier with \texttt{Or} and \texttt{And}, for the two formulations (with/without $\eta$) for the Direct Marketing dataset \cite{moro2014data}. Each classifier is trained and tested on 32 stratified shuffled splits with a 70/30 split (for cross-validation) over the in-sample data (80/20 stratified split). The points correspond to the mean of the balanced accuracy and complexity over those splits. The error bars are given by the standard deviation over the balanced accuracy and complexity on the respective 32 splits for each point. }
  \label{fig:sa_results_and_timing}
\end{figure*}

See \cref{fig:sa_results_and_timing} for the results of an experiment comparing the two formulations (with/without $\eta$) for the \texttt{Or} and \texttt{And} operators. We observe, anecdotally, that the ``without $\eta$'' formulation leads to a similar or better accuracy-complexity curve, and even when both formulations have similar performance, the runtime for the formulation without $\eta$ is much shorter (potentially by orders of magnitude). This is in line with the above theoretical arguments, however we leave a detailed comparison of the TTS for the two formulations for future work. 

Finally, it is worth noting that the elimination of error variables causes the solution space to be biased. The objective function in this case is a sum of squares of violations (residuals) of the sample constraints. Therefore, given two solutions that have the same score $S$ (i.e., degenerate solutions), the solution with the lower sum of squares of violations is preferred. In some sense, one might argue that this bias is reasonable, because the solution that is preferred is ``wrong by less'' than the solution with the larger sum of squares of violations.

\subsection{Number of variables and the nature of the search space}
\label{sec:scaling}

When calculating the Pareto frontier, we plan to solve a series of optimization problems starting from a small complexity bound $m'$ (i.e., \codevar{max\_num\_literals}) and gradually increasing it. For this reason, it is interesting to consider the dependence of the number of variables and the search space size on $m'$. For simplicity, we limit this discussion to non-parametrized operators. 

\begin{figure*}[htb]
  \begin{minipage}{\textwidth}
	\begin{subfigure}[t]{0.48\textwidth}
	        \centering
		\includegraphics[width=\textwidth]{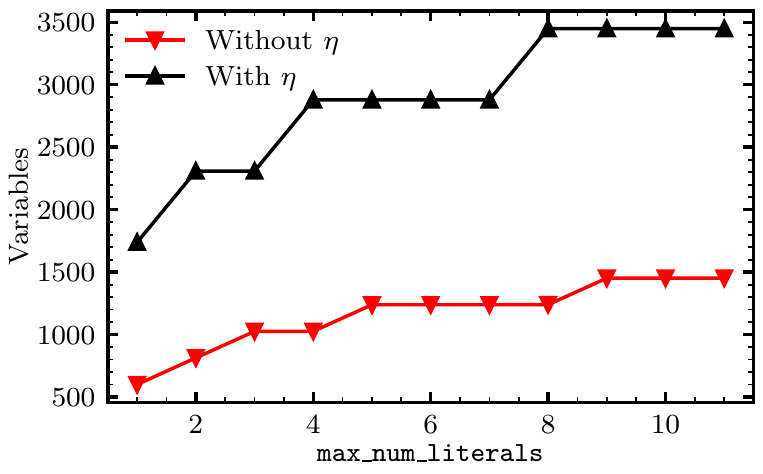}
		\caption{Number of variables}\label{fig:num_variables_breast_cancer}
	\end{subfigure}\hspace{0.5cm}
	\begin{subfigure}[t]{0.48\textwidth}
		\centering
		\includegraphics[width=\textwidth]{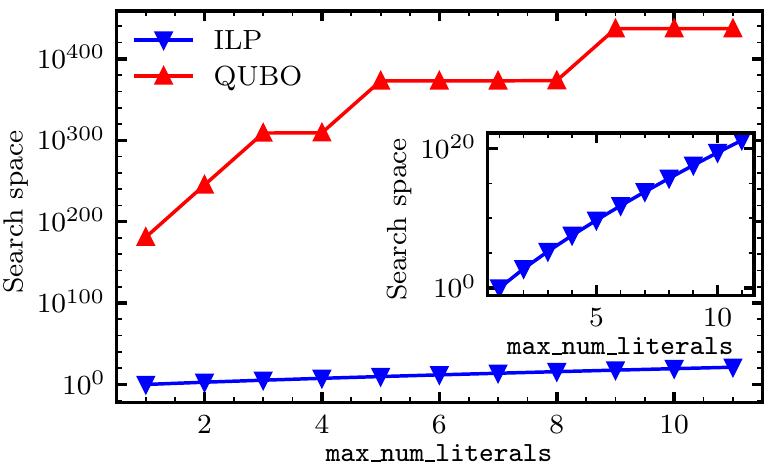}
		\caption{Size of search space}\label{fig:search_space_breast_cancer}
	\end{subfigure}
  \end{minipage}
  \caption{Number of variables and size of search space as a function of the maximum number of literals for the Breast Cancer dataset. The number of variables for an \texttt{Or} rule in the two QUBO formulations, as a function of $m'$ (\codevar{max\_num\_literals}) is plotted in (a)---see \cref{eq:num_variables_or_qubo}. The step-wise form is a result of the binary encoding of the slack in the inequality constraints. The size of the feasible space, which is equal to the size of the search space for the ILP solver, as well as the size of the much larger (feasible and infeasible) space searched by the QUBO formulation (without $\eta$) is plotted in (b). The inset shows a zoomed in version of the former.}
  \label{fig:num_variables_search_space}
\end{figure*}

From \cref{eq:num_variables_or_qubo}, we can see that for a fixed dataset of dimensions $n \times m$, the number of variables in the QUBO formulations increases in a step-wise fashion as $m'$ increases (see \cref{fig:num_variables_breast_cancer}). The search space size for each of the QUBO formulations is given by $2^{\text{\codevar{num\_vars}}}$ where \codevar{num\_vars} is the number of variables in the respective formulation. The size of the search space for the ILP formulation, which is exactly equal to the feasible space, is much smaller and is given by
\begin{equation}
\text{\codevar{num\_feasible}} = \sum_{l=0}^{m'} \binom{2m}{l}.
\end{equation}
For a visual comparison of the sizes of the infeasible space (QUBO) and the feasible space (ILP) see \cref{fig:search_space_breast_cancer}. It is clear that the space searched by the QUBO formulations is far larger than the feasible space, handicapping the QUBO solver. For example, for ${\codevar{max\_num\_literals}\approx 10}$ we find that the QUBO search space surpasses the ILP search space by more than 400 orders of magnitude. This motivates, in part, the usage of the QUBO solver as a subproblem solver, as described later in this paper, instead of for solving the whole problem. When solving smaller subproblems, the size gap between the feasible space and the infeasible space is relatively smaller and might be surmountable by a fast QUBO solver. 

Furthermore, by inspecting the constraints, we can see that moving from a feasible solution to another feasible solution requires flipping more than a single bit. This means that the search space is composed of single feasible solutions, each surrounded by many infeasible solutions with higher objective function values, like islands in an ocean. This situation is typical for constrained QUBO problems. Under these conditions, we expect that a single bit-flip optimizer, such as vanilla simulated annealing, would be at a distinct disadvantage. In contrast, this situation should, in principle, be good for quantum optimization algorithms (including quantum annealing), because these barriers between the feasible solutions are narrow and, therefore, should be relatively easy to tunnel through \cite{farhi2002quantum}.

\section{Benchmarking methodology and results}
\label{sec:results}

In this section we state our research questions, the setup for our numerical experiments, and present and discuss our results.

\subsection{Research questions}

We consider the following research questions (RQ) to demonstrate the effectiveness of our approach: 

\begin{questions}[itemindent=1em]
\item What is the performance of each solution approach with respect to the Pareto frontier, i.e., score vs. complexity? 
\item Does sampling help to scale to large datasets? 
\item Are non-local moves advantageous vs. using just local moves, and under what conditions? 
\end{questions}

\subsection{Benchmarking methodology}

\textbf{Datasets} --
See \cref{tab:datasets} for the binary classification datasets included in our experiments. We have selected datasets with varied characteristics. In particular, the number of samples ranges from 195 (very small) to ${\mathord\sim1.3\cdot 10^5}$ (very large), and the number of binarized features ranges from 67 to 300. 

Several datasets included in our study contain samples with missing data. Those samples were removed prior to using the data. We removed 393 samples from the Airline Customer Satisfaction dataset, 11 samples from the Customer Churn dataset, and 2596 samples from the Home Equity Default dataset. The first two are negligible, but the latter comprise of 44\% of the data. 

\begin{table*}[htb]
	\centering
 	\caption{Datasets included in our experiments. ``Rows'' is the number of data samples with no missing values, ``Features'' is the number of features provided, ``Binarized'' is the number of features after binarization, ``Majority'' is the fraction of data rows belonging to the larger of the two classes, and ``Ref.'' is a reference for each dataset. \label{tab:datasets}}
	\begin{tabular}{@{}lrrrrr@{}}
		\toprule
		Name & Rows & Features & Binarized & Majority & Ref. \\
		\midrule
		Airline Customer Satisfaction & 129,487 & 22 & 119 & 0.55 & \cite{kaggleairlinecustomersatisfaction} \\
		Breast Cancer  & 569 & 30 & 300 & 0.63 & \cite{wolberg1992breast} \\ 
		Credit Card Default & 30,000 & 23 & 185 & 0.78 & \cite{yeh2009comparisons} \\
		Credit Risk & 1,000 & 20 & 92 & 0.70 & \cite{dua2019} \\
		Customer Churn & 7,043 & 19 & 67 & 0.73 & \cite{kaggletelco} \\
		Direct Marketing & 41,188 & 20 & 124 & 0.89 & \cite{moro2014data} \\
		Home Equity Default & 3,364 & 12 & 93 & 0.80 & \cite{kagglehomeequity} \\
		Online Shoppers Intention & 12,330 & 17 & 104 & 0.85 & \cite{sakar2019real} \\
		Parkinsons         & 195 & 22 & 217 & 0.75 & \cite{little2007exploiting} \\
		\bottomrule
	\end{tabular}
\end{table*}

\textbf{Binarization} --
In order to form Boolean rules, all the input features must be binary. For this reason, we binarize any features that are not already binary. Features that are already binary do not require special treatment. Features that are categorical, or numerical with few unique values, are binarized using a one-hot encoding. For features that are numerical with many unique values, many binarization methods are available, including splitting them into equal-count bins, into equal-width bins, and into bins that maximize the information gain \cite{fayyad1993multi}. We hypothesize that the choice of method of binarization can have a strong effect on downstream ML models, though we leave this for future work. 

In this paper, the binarization is carried out by binning the features followed by encoding the features using a one-hot encoding. The binning is carried out by calculating \codevar{num\_bins} quantiles for each feature. For each quantile, a single ``up'' bin is defined, extending from that quantile value to infinity. Because our classifiers all include the negated features, the corresponding ``down'' bins are already included by way of those negated features. In all experiments we set \codevar{num\_bins}~$=10$.\\

\textbf{Classifiers} --
We include several classifiers in our experiments. First, the baseline classifiers:
\begin{itemize}
\item \textit{Most frequent} -- A naive classifier that always outputs the label that is most frequent in the training data. This classifier always gives exactly 0.5 for balanced accuracy. We exclude this classifier from the figures to reduce clutter and because we consider it as a lower baseline. This classifier is easily outperformed by all the other classifiers. 
\item \textit{Single feature} -- A classifier that consists of a simple rule, containing just a single feature. The rule is determined in training by exhaustively checking all possible rules consisting of a single feature or a single negated feature. 
\item \textit{Decision tree} -- A decision tree classifier, as implemented in \textsc{scikit-learn} 
\cite{scikit-learn} with {\codevar{class\_weight}~$=$~``balanced''}. Note that decision trees are able to take non-binary inputs, so we also include results obtained by training a decision tree on the raw data, with no binarization. In order to control the complexity of the decision tree, we vary \codevar{max\_depth} which sets the maximum depth of the trained decision tree. The complexity is given by the number of split nodes (as described in \cref{sec:motivation}). 
\end{itemize}

Then, the depth-one QUBO and ILP classifiers:
\begin{itemize}
\item \textit{ILP rule} -- A classifier that solves the ILP formulations described in \crefrange{sec:ilp_or}{sec:ilp_parameterized_operators} for a depth-one rule with a given operator at the root, utilizing \textsc{FICO Xpress} (version 9.0.1) with a timeout of one hour. In order to limit the size of the problems, a maximum of 3,000 samples is used for each cross-validation split. 

\item \textit{QUBO rule} -- A classifier that solves the QUBO formulations described in \cref{sec:qubo_formulations} for a depth-one rule with a given operator at the root, utilizing simulated annealing as implemented in \textsc{dwave-neal} with {\codevar{num\_reads}~$=100$}, and {\codevar{num\_sweeps}~$=2000$}. In order to limit the size of the problems, a maximum of 3,000 samples is used for each cross-validation split. The results were generally worse than the ILP classifier results (and are guaranteed not to be better in score), so to reduce clutter we do not include them below (but see some QUBO results in \cref{fig:sa_results_and_timing}). A likely explanation for the underwhelming results is explained in \cref{sec:scaling} - this is a single bit-flip optimizer, but going from feasible solution to feasible solution in our QUBO formulations requires flipping more than one variable at a time. 
\end{itemize}

Finally, the native local solvers:
\begin{itemize} 
\item \textit{SA native local rule} -- The simulated annealing native local rule classifier described in \cref{sec:local_solver}, with  ${\codevar{num\_starts}~=20}$, and ${\codevar{num\_iterations}~=2000}$. The temperatures follow a geometric schedule from 0.2 to $10^{-6}$.
\item \textit{SA native non-local rule} -- The simulated annealing native local rule classifier, with additional ILP-powered non-local moves as described in \cref{sec:nonlocal_moves}. Uses the same parameter values as the native local rule classifier, and in addition the burn-in period consists of the first third of the steps, ${\codevar{patience}~=10}$, ${\codevar{max\_samples}~=100}$, and the timeout for the ILP solver was set to one second. 
\end{itemize}

\textbf{Cross-validation} --
Each dataset is split into in-sample data and out-of-sample data using an 80/20 stratified split. The in-sample data are then shuffled and split 32 times (unless indicated otherwise in each experiment) into train/test data with a 70/30 stratified split. All benchmarking runs are done on Amazon EC2 utilizing \texttt{c5a.16xlarge} instances, which have 64 vCPUs and 128 GiB of RAM. Cross-validation for each classifier and dataset is generally performed on 32 splits in parallel in separate processes, on the same instance.

\textbf{Hyperparameter optimization} --
Very minimal hyperparameter optimization is performed---it is assumed that the results could be improved by parameter tuning, which is not the focus of this work. In addition, it is possible that using advanced techniques for solving ILPs, such as column generation, would improve the results measurably. Note that ${\lambda=0}$ is used in all experiments to simplify the analysis.

\subsection{Results and Discussion}

\textbf{Pareto frontier (RQ1)} --
On each dataset, for each classifier, we report the mean and standard deviation of the balanced accuracy and of the complexity over those splits, for the train/test portions of the data. Some classifiers are unparameterized, so we report a single point for each dataset (single feature classifier and most frequent classifier). The decision tree classifier results are obtained by changing the \codevar{max\_depth} parameter over a range of values. For the depth-one QUBO and ILP classifiers, the results are obtained by changing the \codevar{max\_num\_literals} parameter over a range of values. Finally, for the native local optimizers the results are obtained by changing the \codevar{max\_complexity} parameter over a range of values. The decision tree is also run on the raw data, without a binarizer, which is indicated by the suffix ``(NB)''. 

For all datasets except the three smallest ones (up to 1,000 samples), the train and test results are virtually indistinguishable. For this reason, we present train and test results for the smaller datasets, but only test results for the larger datasets. The near identity of the train and test results is likely explained by the fact that our classifiers are (by design) highly regularized by virtue of the binarization and their low complexity. Therefore, they do not seem to be able to overfit the data, as long as the data size is not very small. 

See \cref{fig:results_native_local_solver} for train and test results for the three smallest datasets, and \cref{fig:results_native_local_solver2} for test results for the six larger datasets. Finally, example rules found by the native local optimizer for each of the datasets are presented in \cref{tab:example_rules_native_local_solver}.

\begin{table*}[htb]
\footnotesize
	\centering
 	\caption{Example rules obtained by the native local solver for each dataset. ``Dataset'' is the name of the dataset, and ``Rule'' is the best rule found by the first of the cross-validation splits, for the case \codevar{max\_complexity}~$=4$. The variable names in the rules are as obtained from the original datasets. Negations of features are post-processed out by reversing the relationship in the respective feature name (e.g., $\mathord\sim a=3 \to a \neq 3$).}\label{tab:example_rules_native_local_solver}
	\begin{tabular}{@{}ll@{}} 
		\toprule
		Dataset & Rule \\
		\midrule
                Airline Customer Satisfaction & \texttt{And(}Inflight entertainment~$\neq$~5, Inflight entertainment~$\neq$~4, Seat comfort~$\neq$~0\texttt{)} \\
                Breast Cancer & \texttt{AtMost1(}worst concave points~$\leq$~0.1533, worst radius~$\leq$~16.43, mean texture~$\leq$~15.3036\texttt{)} \\
                Credit Card Default & \texttt{Or(}PAY\_2~$>$~0, PAY\_0~$>$~0, PAY\_4~$>$~0\texttt{)} \\
                Credit Risk & \texttt{Choose1(}checking\_status~$=$~no checking, checking\_status~$<$~200, property\_magnitude~$=$~real estate\texttt{)} \\
                Customer Churn & \texttt{AtMost1(}tenure~$>$~5, Contract~$\neq$~Month-to-month, InternetService~$\neq$~Fiber optic\texttt{)} \\
                Direct Marketing & \texttt{Or(}duration~$>$~393, nr.employed~$\leq$~5076.2, month~$=$~mar\texttt{)} \\
                Home Equity Default & \texttt{Or(}DEBTINC~$>$~41.6283, DELINQ~$\neq$~0.0, CLNO~$\leq$~11\texttt{)} \\
                Online Shoppers Intention & \texttt{AtMost1(}PageValues~$\leq$~5.5514, PageValues~$\leq$~0, BounceRates~$>$~0.025\texttt{)} \\
                Parkinsons & \texttt{AtMost1(}spread1~$>$~-6.3025, spread2~$>$~0.1995, Jitter:DDP~$>$~0.0059\texttt{)} \\                \bottomrule
	\end{tabular}
\end{table*}

\begin{figure*}[htb]
  	\begin{subfigure}[t]{1.0 \textwidth}
		\centering
		\includegraphics[width=1.0 \textwidth]{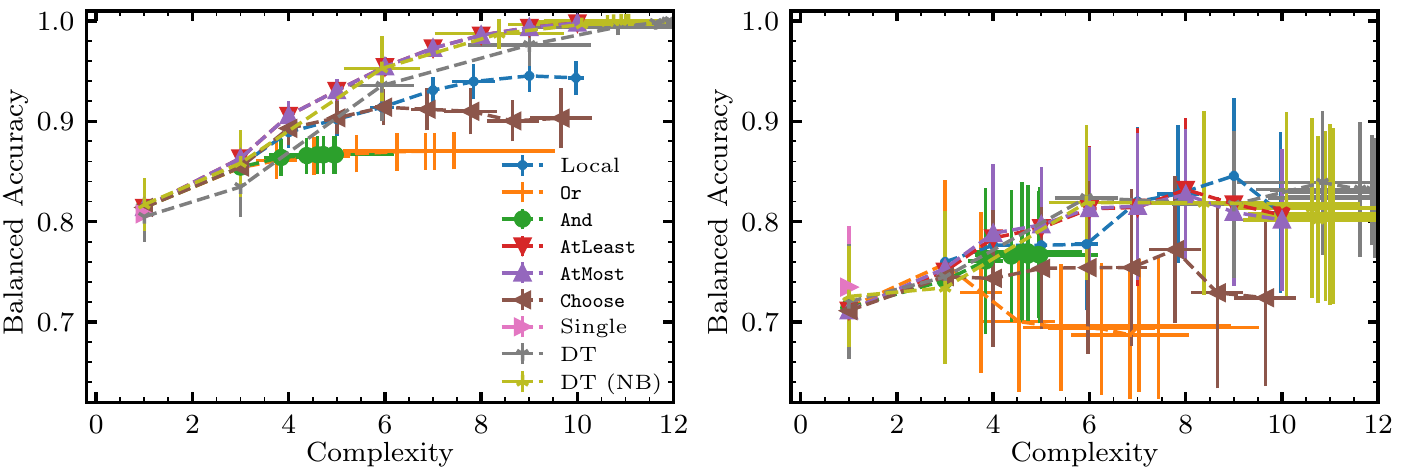}
		\caption{Parkinsons}\label{fig:native_local_results_parkinsons}
	\end{subfigure}\\
	\begin{subfigure}[t]{1.0 \textwidth}
		\centering
		\includegraphics[width=1.0 \textwidth]{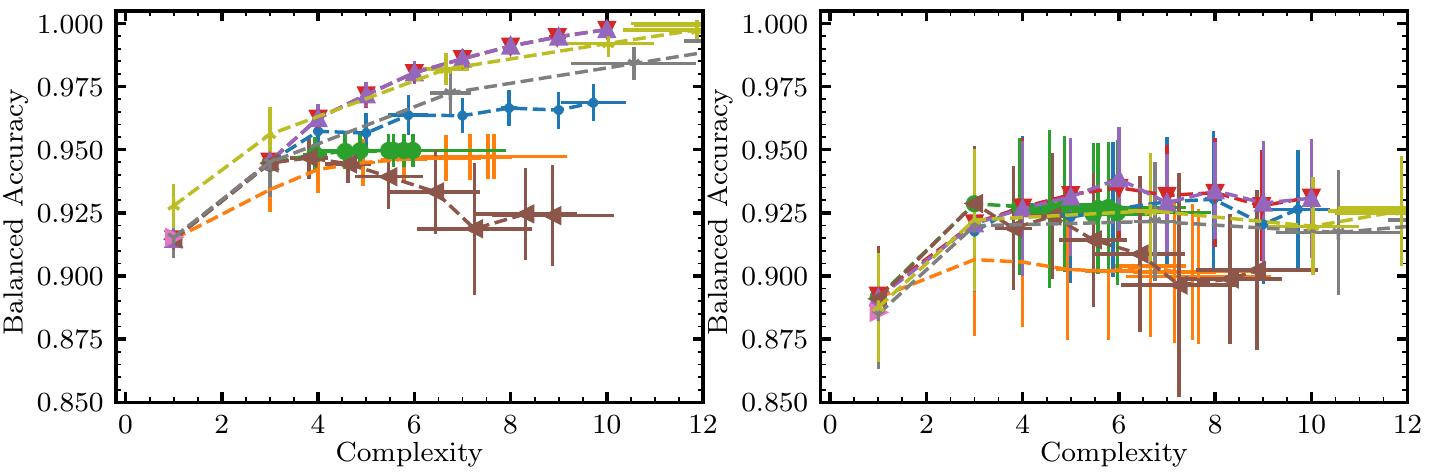}
		\caption{Breast Cancer}\label{fig:native_local_results_breast_cancer}
	\end{subfigure}\\
	\begin{subfigure}[t]{1.0 \textwidth}
		\centering
		\includegraphics[width=1.0 \textwidth]{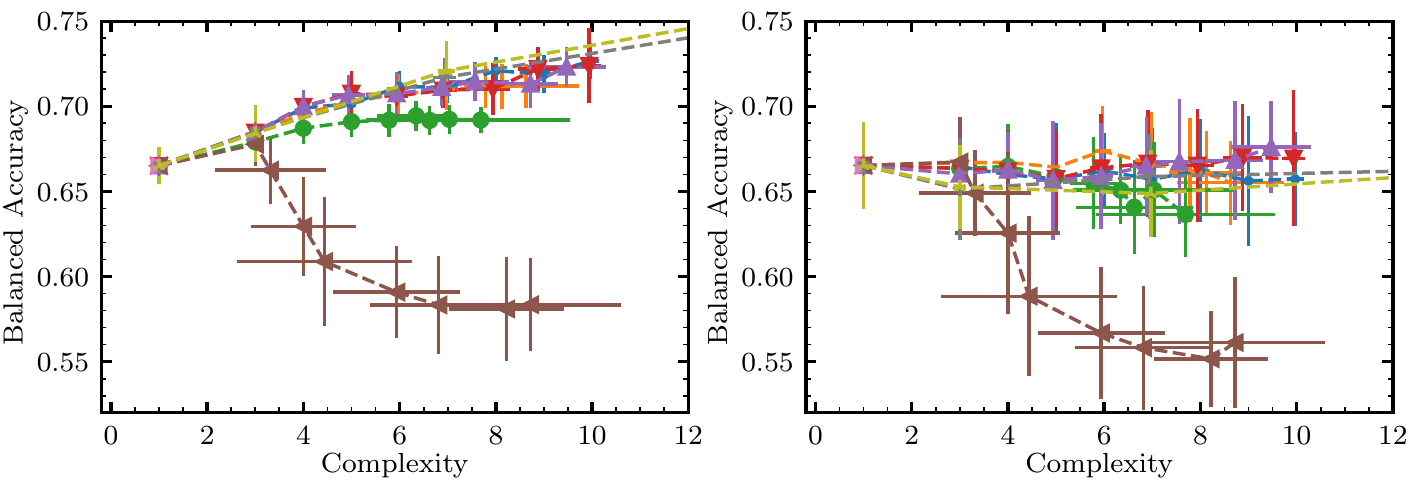}
		\caption{Credit Risk}\label{fig:native_local_results_credit_risk}
	\end{subfigure}
  \caption{Train (left) and test (right) results for the native local solver (``Local'') vs. the depth-one ILP classifiers (indicated by the name of the respective operator, e.g., \texttt{Or}), the single feature classifier (``Single''), the decision tree (``DT''), and the decision tree with no binarizer (``DT (NB)''), for the three smallest datasets. Each classifier is trained and tested on 32 stratified shuffled splits with a 70/30 split (for cross-validation) over the in-sample data (80/20 stratified split). The points correspond to the mean of the balanced accuracy and complexity over those splits. The error bars are given by the standard deviation over the balanced accuracy and complexity on the respective 32 splits for each point. Continued in \cref{fig:results_native_local_solver2}.}\label{fig:results_native_local_solver}
\end{figure*}

\begin{figure*}[htb]
  	\begin{subfigure}[t]{0.49 \textwidth}
		\centering
		\includegraphics[width=1.0 \textwidth]{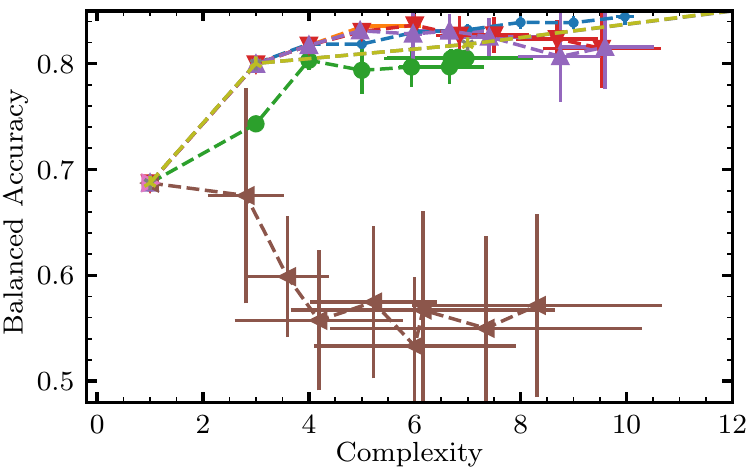}
		\caption{Airline Customer Satisfaction}\label{fig:native_local_results_airline_customer_satisfaction}
	\end{subfigure}
    \hspace{0.1cm}
   	\begin{subfigure}[t]{0.49 \textwidth}
		\centering
		\includegraphics[width=1.0 \textwidth]{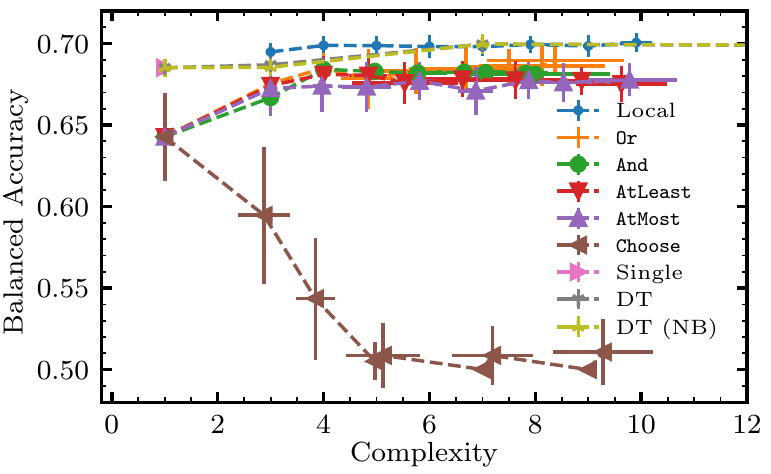}
		\caption{Credit Card Default}\label{fig:native_local_results_credit_card_default}
	\end{subfigure}\\
	\begin{subfigure}[t]{0.49 \textwidth}
		\centering
		\includegraphics[width=1.0 \textwidth]{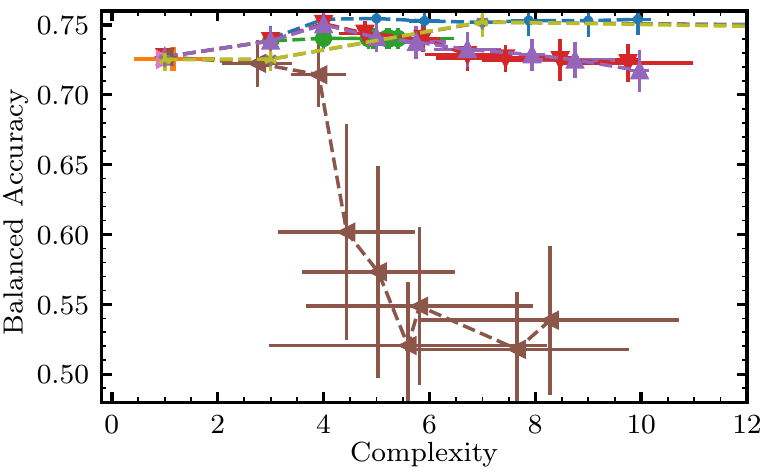}
		\caption{Customer Churn}\label{fig:native_local_results_credit_customer_churn}
    \end{subfigure}
    \hspace{0.1cm}
	\begin{subfigure}[t]{0.49 \textwidth}
		\centering
		\includegraphics[width=1.0 \textwidth]{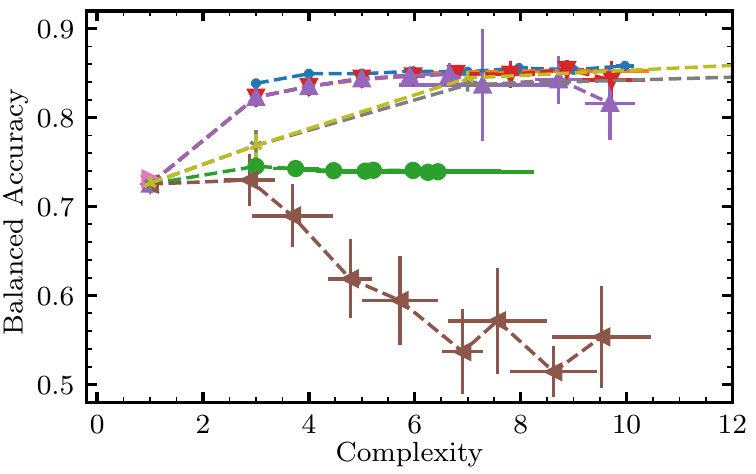}
		\caption{Direct Marketing}\label{fig:native_local_results_credit_direct_marketing}
    \end{subfigure}\\
	\begin{subfigure}[t]{0.49 \textwidth}
		\centering
		\includegraphics[width=1.0 \textwidth]{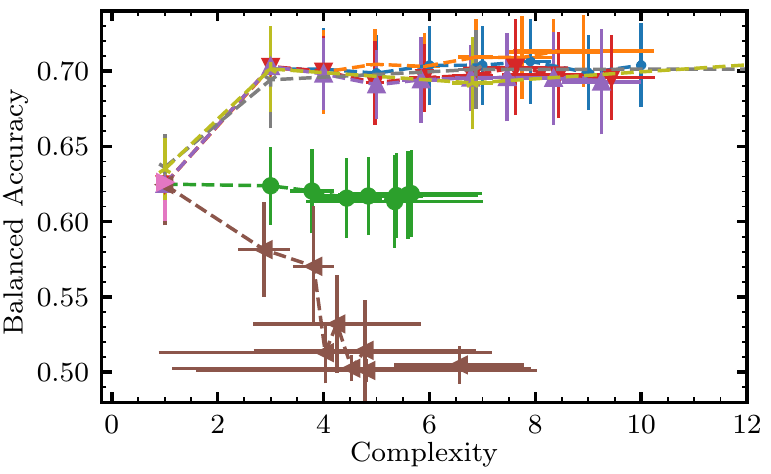}
		\caption{Home Equity Default}\label{fig:native_local_results_home_equity_default}
    \end{subfigure}
    \hspace{0.1cm}
  	\begin{subfigure}[t]{0.49 \textwidth}
		\centering
		\includegraphics[width=1.0 \textwidth]{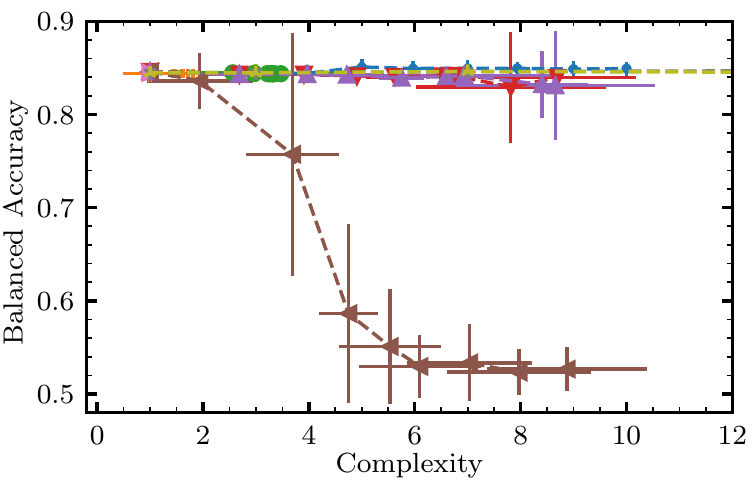}
		\caption{Online Shoppers Intention}\label{fig:native_local_results_online_shoppers_intention}
	\end{subfigure}
  \caption{Continued from \cref{fig:results_native_local_solver}. Test results for the six largest datasets. For these larger datasets the train and test results are virtually indistinguishable, likely due to the highly regularized models used. For this reason only the test results are presented.}\label{fig:results_native_local_solver2}
\end{figure*}

The native local rule classifier is found to be generally competitive with the other classifiers; it achieves similar scores to the decision tree, despite the decision tree involving a more flexible optimization procedure. The ILP classifiers are shallower, but still achieve similar scores to the native local solver for most datasets. However, the ILP classifiers require significantly longer run times. This suggests that the additional depth does not always provide an advantage. However, for some of the datasets, such as Direct Marketing and Customer Churn, there is a slight advantage to the native local solver, and hence an apparent advantage to deeper rules. However, note that deeper rules could be considered less interpretable, and so less desirable for the XAI use case.  

Of note, the single feature rule classifier baseline beats or matches all other classifiers on the Credit Card Default, Credit Risk, and Online Shoppers Intention datasets, suggesting  that for those datasets a more complex model than those included in this work is required in order to achieve higher scores. The best score achieved varies significantly across the datasets studied, suggesting that they feature a range of hardnesses. 

For all datasets except Parkinsons, the great majority of the decision tree classifier's results have such a high complexity that they are outside of the plot. This is another example of the issue pointed out in \cref{sec:motivation}---decision trees tend to yield high complexity trees/rules even for single-digit values of \codevar{max\_depth}.

Comparing the different ILP operators, the results suggest that the parametrized operators \texttt{AtMost} and \texttt{AtLeast} offer an advantage over the unparameterized operators, which could be explained by their additional expressivity. We note that performance of \texttt{AtMost} and \texttt{AtLeast} is largely identical, in line with the similarity of these operators. The results for the third parametrized operator, \texttt{Choose}, are poor for some of the datasets, which is likely due to the larger problem size in this formulation, thereby resulting in the ILP solver timing out before finding a good solution. In fact, many of the other ILP runs timed out despite sampling down the larger datasets to 3,000 samples (for training), which suggests that trying to prove optimality for large datasets might not be scalable. Furthermore, as anticipated in \cref{sec:ilp_and}, for some of the datasets the results for the \texttt{And} operator are better than the results for the \texttt{Or} operator (for example, for the Breast Cancer dataset).

\begin{figure*}[htb]
  	\begin{subfigure}[t]{1.0 \textwidth}
		\centering
		\includegraphics[width=1.0 \textwidth]{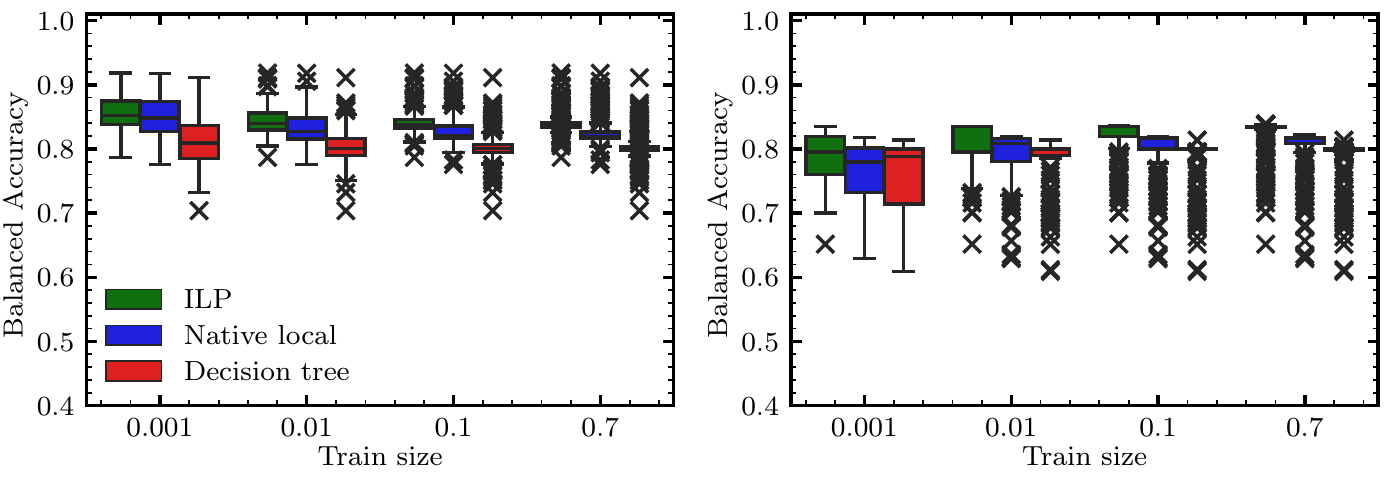}
		\caption{Airline Customer Satisfaction}\label{fig:sampling_airline_customer_satisfaction}
	\end{subfigure}\\
	\begin{subfigure}[t]{1.0 \textwidth}
		\centering
		\includegraphics[width=1.0 \textwidth]{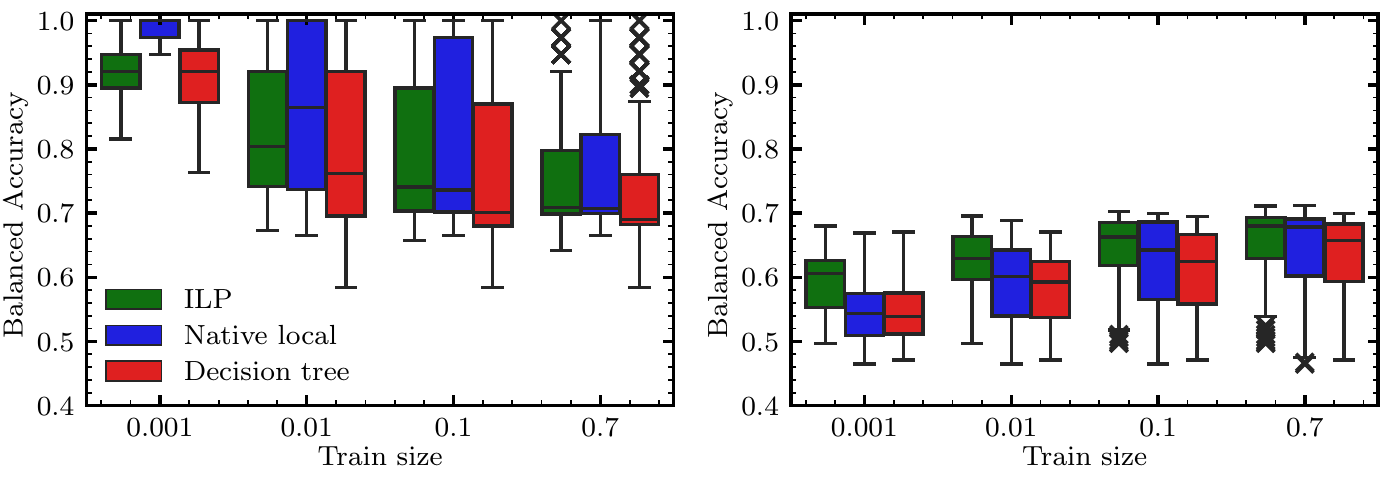}
		\caption{Credit Card Default}\label{fig:sampling_credit_card_default}
	\end{subfigure}\\
	\begin{subfigure}[t]{1.0 \textwidth}
		\centering
		\includegraphics[width=1.0 \textwidth]{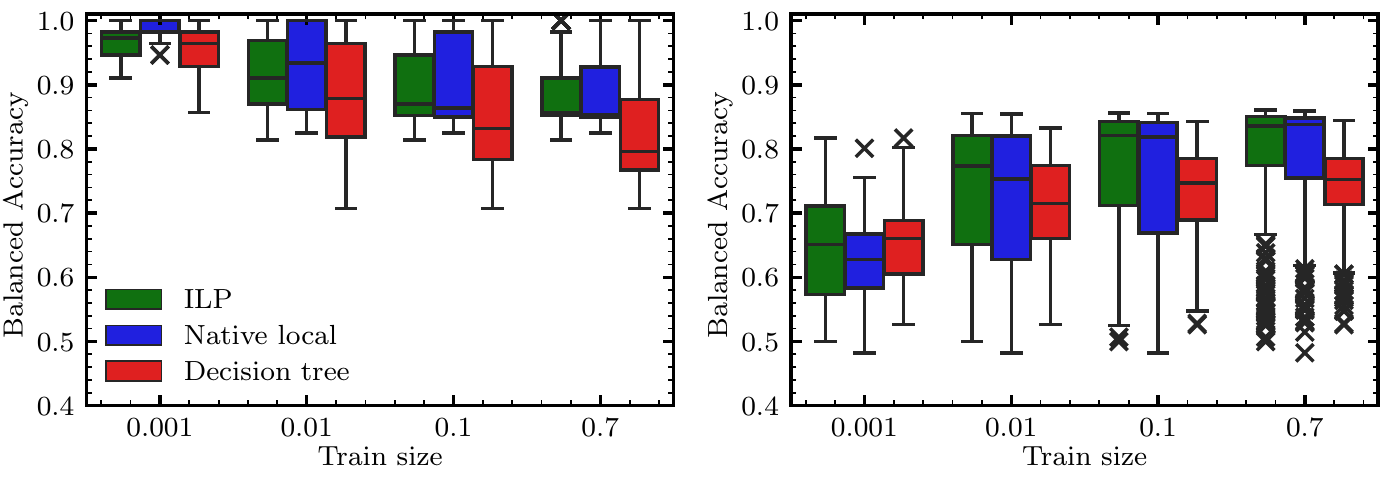}
		\caption{Direct Marketing}\label{fig:sampling_direct_marketing}
	\end{subfigure}
  \caption{Train (left) and test (right) results for various classifiers as a function of the percent of the in-sample data that is included in the training. The ILP classifier uses the \texttt{Or} operator and \codevar{max\_num\_literals}~$=4$. The native local solver uses \codevar{max\_complexity}~$=5$, and the decision tree uses \codevar{max\_depth}~$=2$. For each dataset, the classifiers are trained and tested on 64 stratified shuffled splits (for cross-validation) of the in-sample data (from an 80/20 stratified split) with the indicated train size, and the rest of the in-sample data are used as the validation set. The box plots show the balanced accuracy for each sample size for each of the datasets.}\label{fig:results_sampling}
\end{figure*}

\textbf{Sampling (RQ2)} --
The main objective of this experiment is to assess whether large datasets can be tackled via sampling, i.e., by selecting a subsample of the data to be used for training the classifier (see \cref{fig:results_sampling}). This experiment is run only on the three largest datasets to allow for enough room to change the sample size. 

We observe that the training score goes down with train-sample size until saturation, presumably since it is harder to fit a larger dataset. At the same time, the test score goes up with train-sample size, presumably since the larger sample provides better representation and, therefore, the ability to generalize. The score distribution on the train and test sets generally narrows down with increasing train-sample size, presumably converging on the population's score. We observe that the train and test scores are initially very different for small train-sample sizes but converge to similar values for large train-sample sizes. These results suggest that several thousand samples might be enough in order to reach a stable score for the ILP and native local solvers, in comparison to Deep Learning techniques that often require far more data (for example \cite{van2014modern}). We comment also that the sampling procedure is attractive due to its simplicity and due to it being classifier agnostic.

\begin{figure*}[htb]
  	\begin{subfigure}[t]{1.0 \textwidth}
		\centering
		\includegraphics[width=1.0 \textwidth]{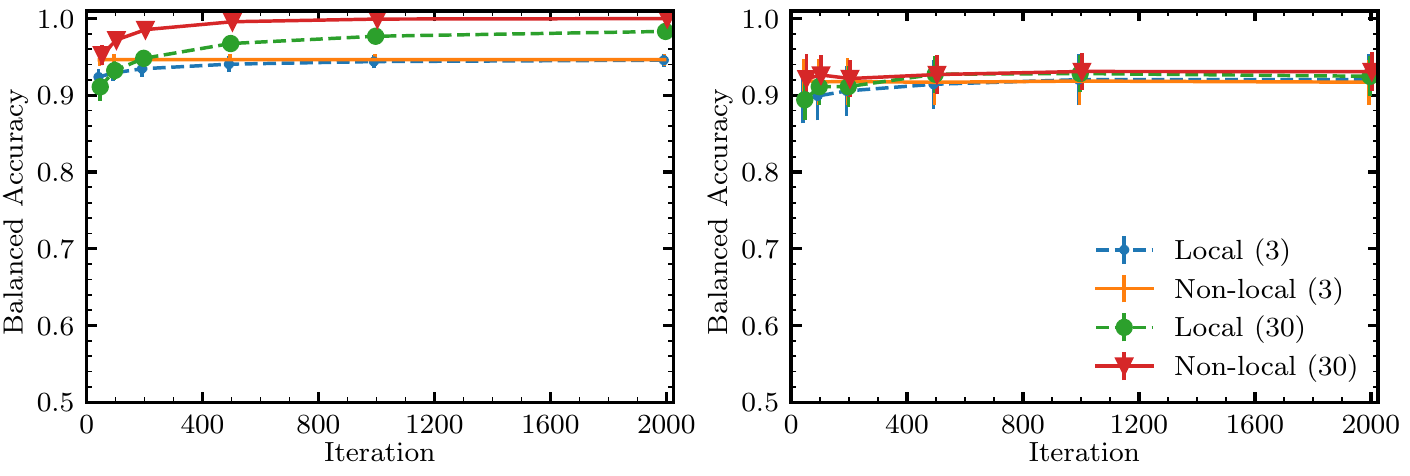}
		\caption{Breast Cancer}\label{fig:native_non_local_breast_cancer}
	\end{subfigure}\\
	\begin{subfigure}[t]{1.0 \textwidth}
		\centering
		\includegraphics[width=1.0 \textwidth]{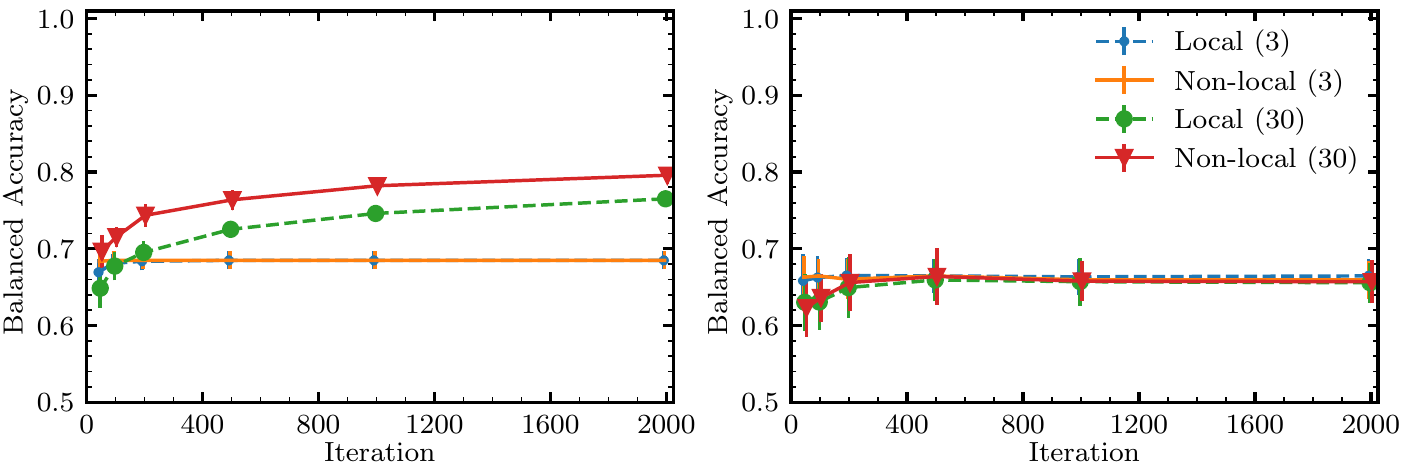}
		\caption{Credit Risk}\label{fig:native_non_local_credit_risk}
	\end{subfigure}\\
	\begin{subfigure}[t]{1.0 \textwidth}
		\centering
		\includegraphics[width=1.0 \textwidth]{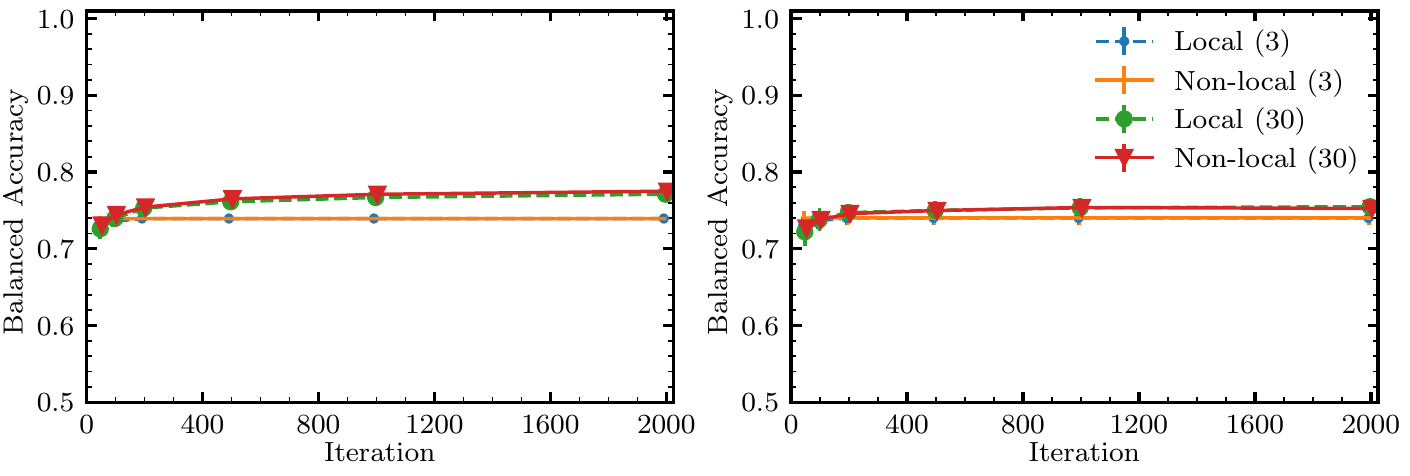}
		\caption{Customer Churn}\label{fig:native_non_local_customer_churn}
	\end{subfigure}
  \caption{Train (left) and test (right) results for the native local solver and native local solver with non-local moves, as a function of the number of iterations. For each dataset, the classifiers are trained and tested on 32 stratified shuffled splits of the in-sample data (from an 80/20 stratified split) with the indicated train size, and the rest of the in-sample data are used as the validation set. Jitter added to aid in viewing.}\label{fig:results_native_non_local}
\end{figure*}

\textbf{Native non-local solver (RQ3)} --
The main objective of this experiment is to assess whether the non-local moves provide an advantage and under which conditions. With this in mind we vary both \codevar{max\_complexity} and \codevar{num\_iterations} in \cref{fig:results_native_non_local}.

In the train results, it is clear that the non-local moves provide a meaningful improvement (several percent in Balanced Accuracy) for two out of the three datasets (and a marginal improvement for the third), but only at the higher complexity, suggesting that one can use the non-local moves to fit the data with a smaller number of iterations. If a solver is available that can solve the optimization problem to find non-local moves fast, then using such a solver might lead to faster training. At lower complexity, there is not enough ``room'' for the non-local moves to operate, so they do not provide an advantage. 

In the test results, the error bars are overlapping, so it is not possible to make a strong statement. However, we note that the point estimates for the mean are slightly improved over the whole range of complexities for all three datasets.

Note that the very short timeout for the ILP solver to find each non-local move likely curtailed the solver's ability to find good moves. In addition, ILP solvers typically have many parameters that control their operation, which were not adjusted in this case. It is likely that asking the solver to focus on quickly finding good solutions would improve the results. We leave this direction for future research.  

\clearpage 

\section{Conclusions and Outlook}
\label{sec:conclusions}

We have proposed a class of interpretable ML classification models based on expressive Boolean formulas, and discussed in detail the underlying training procedure based on native local optimization techniques, with the optional addition of non-local moves. These non-local moves are proposed by solving an optimization problem that can be formulated as an ILP or a QUBO problem. As such, we foresee an opportunity to gain potential speedups by using hardware accelerators, including future quantum computers. Solving the XAI problem to optimality is not strictly required in most practical scenarios, thus potentially lowering the requirements on noise for quantum devices. The formulations we have introduced are usable also as standalone classifiers in which the Boolean rule forming the basis for the classifier is very shallow (depth-one). Future work may center on applying these classifiers to other datasets, introducing new operators, or applying these concepts to other uses cases. 

\begin{acknowledgments}

This work is a collaboration between Fidelity Center for Applied Technology, Fidelity Labs, LLC., and Amazon Quantum Solutions Lab. The authors would like to thank Cece Brooks, Michael Dascal, Cory Thigpen, Ed Cady, Kyle Booth, and Thomas H\"{a}ner for fruitful discussions. The Fidelity publishing approval number for this paper is 1084542.1.0. 

\end{acknowledgments}

\bibliographystyle{unsrtnat}
\bibliography{refs}

\end{document}